\documentclass[format=acmsmall, review=false, screen=true]{acmart}

\usepackage{amsmath}
\usepackage{graphicx}

\usepackage{graphicx,calc}
\newlength\myheight
\newlength\mydepth
\settototalheight\myheight{Xygp}
\newcommand*\inlinegraphics[1]{
\settototalheight\myheight{Xygp}
\settodepth\mydepth{Xygp}
\raisebox{-\mydepth}{\includegraphics[height=\myheight]{#1}}
}

\begin{document}
\title{A Classifier Using Global Character Level and Local Sub-unit Level Features for Hindi Online Handwritten Character Recognition}

\author{Anand Sharma}
\email{anand.sharma@miet.ac.in}
\affiliation{
\institution{MIET}
\city{Meerut}
\country{India}}
\author{A. G. Ramakrishnan}
\email{agr@iisc.ac.in}
\affiliation{
\institution{IISc}
\city{Bangaluru}
\country{India}}

\begin{abstract}
\indent A classifier is developed that defines a joint distribution of global character features, number of sub-units and local sub-unit features to model Hindi online handwritten characters. The classifier uses latent variables to model the structure of sub-units. The classifier uses histograms of points, orientations, and dynamics of orientations (HPOD) features to represent characters at global character level and local sub-unit level and is independent of character stroke order and stroke direction variations. The parameters of the classifier is estimated using maximum likelihood method.\\
\indent Different classifiers and features used in other studies are considered in this study for classification performance comparison with the developed classifier. The classifiers considered are Second Order Statistics (SOS), Sub-space (SS), Fisher Discriminant (FD), Feedforward Neural Network (FFN) and Support Vector Machines (SVM) and the features considered are Spatio Temporal (ST), Discrete Fourier Transform (DFT), Discrete Cosine Transform (SCT), Discrete Wavelet Transform (DWT), Spatial (SP) and Histograms of Oriented Gradients (HOG).\\
\indent Hindi character datasets used for training and testing the developed classifier consist of samples of handwritten characters from 96 different character classes. There are 12832 samples with an average of 133 samples per character class in the training set and 2821 samples with an average of 29 samples per character class in the testing set.\\ 
\indent The developed classifier has the highest accuracy of 93.5\% on the testing set compared to that of the classifiers trained on different features extracted from the same training set and evaluated on the same testing set considered in this study.
\end{abstract}



\maketitle

\section{Introduction}
\indent Interaction with computers by inputting characters from large Hindi character set using Latin script based keyboard is difficult. A computer having pen computing environment and online handwriting recognition capability can facilitate easy interaction by allowing an individual to give handwritten input to the computer in Hindi script. Online handwriting recognition system recognizes handwriting produced by the pen movement on a computer interface by analysing the structure of the handwriting and producing the corresponding script unicodes.\\
\indent Researchers studying online handwriting recognition have used different types of classifiers for capturing variations and structures in samples of online handwritten characters for accurate character recognition. Aparna et al. \cite{clse} use string matching approach for identification of strokes having shape based representation and recognize a character by identifying all the component strokes. In the studies done by Sharma et al. \cite{clsd} and Joshi et al. \cite{clsa}, online handwritten characters are recognized using elastic matching technique. HMM model is used for character recognition by Parui et al. \cite{clsc} and Belhe et al. \cite{troaw}. Sub-space based method is used by Prasad et al. \cite{clsb} and Joshi et al.\cite{troao} for character recognition. Combination of HMM and NN models are used for classification of characters by Connell et al. \cite{troan}. SVM has been used for stroke classification by Swethalakshmi et al. \cite{troaq}. SOS based classifier has been used by Lajish et al. \cite{troau} for stroke classification. FNN and convolutional neural network have been used by Kubatur et al. \cite{troav} and Mehrotra et al. \cite{troax}, respectively, for character recognition. The parameters of all these classifiers are determined by training them with the samples of online handwritten characters.\\ 
\indent Classifiers used in other studies that have been considered in this study for performance comparison with the developed classifier are SOS, SS, FD, FNN and SVM classifiers. These classifiers are trained with ST, DFT, DCT, DWT, SP, and HOG features. The classifiers that use ST, DFT, DCT and DWT features are dependent on the stroke direction and order variations. The classifiers that use SP and HOG features are independent of such variations but, like the aforementioned features, are global in nature and do not capture local variations, Sharma \cite{hpod}.\\
\indent The classifier developed in this study uses HPOD features developed by Sharma \cite{hpod} and represents characters at global character level and local sub-unit level. This classifier is independent of stroke direction and order variations and captures both the global and local variations in the characters' structure. The parameters of the classifier are determined by maximum likelihood method. \\
\indent This paper is organized into five Sections. Section 2 describes Hindi character set and Hindi character dataset. Different classifiers along with the developed classifier are explained in Section 3. Experiments and results are given in Section 4. Section 5 summarizes the contribution of this work. 

\section{Hindi characters}
\indent Hindi language is written using Devanagari script \cite{troiscii} and so Hindi characters are subset of Devanagari characters. Online handwritten samples of some of the the characters used for writing Hindi language have been collected for training the developed classifier and for comparing the performance of the developed classifier with that of the classifiers considered in this study.
\subsection{Hindi character set}
\indent Hindi characters considered for character classification are vowels, consonants, pure consonants, nasalization sign, vowel omission sign, vowel signs, consonant with vowel sign, and conjuncts. Strokes with different shapes are also considered in addition to these characters. Figure 1 shows subset of Hindi vowels considered for analysis. The other vowels can be obtained by a combination of these vowels and one or more vowel signs. For example, the vowel {\inlinegraphics{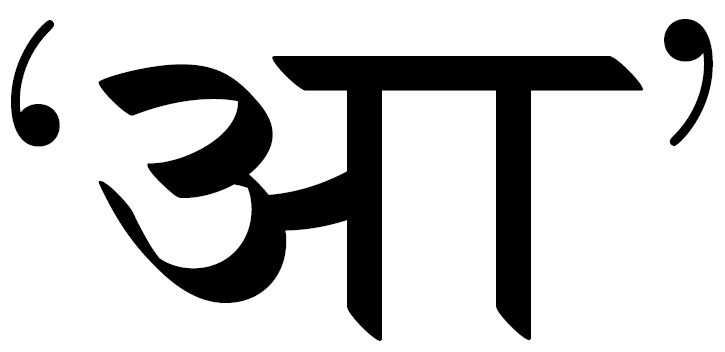}} can be obtained as a combination of the symbol 1 in Fig. 1 and the symbol 67 in Fig. 4.
\begin{figure}[ht]
\begin{center}
$\begin{array}{cccccc}
\includegraphics[width=0.4 in]{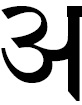}&
\includegraphics[width=0.4 in]{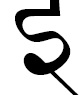}&
\includegraphics[width=0.4 in]{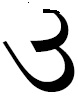}&
\includegraphics[width=0.5 in]{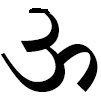}&
\includegraphics[width=0.6 in]{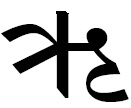}&
\includegraphics[width=0.35 in]{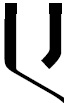}\\
1&2&3&4&5&6\\
\end{array}$
\caption{Basic Hindi vowels used as recognition primitives.}
\end{center}
\end{figure}
All the consonants have an implicit vowel `a' \cite{troiscii}. Figure 2  shows the subset of Hindi consonants considered.
\begin{figure}[ht]
\begin{center}
$\begin{array}{cccccccccc}
\includegraphics[width=0.45 in]{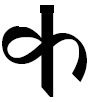}&
\includegraphics[width=0.45 in]{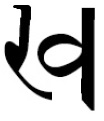}&
\includegraphics[width=0.4 in]{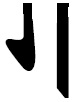}&
\includegraphics[width=0.4 in]{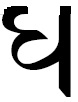}&
\includegraphics[width=0.4 in]{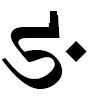}&
\includegraphics[width=0.4 in]{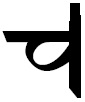}&
\includegraphics[width=0.4 in]{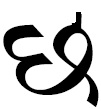}&
\includegraphics[width=0.4 in]{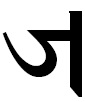}&
\includegraphics[width=0.4 in]{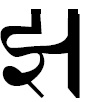}&
\includegraphics[width=0.4 in]{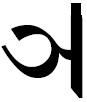}\\
7&8&9&10&11&12&13&14&15&16\\
\includegraphics[width=0.4 in]{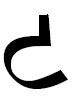}&
\includegraphics[width=0.4 in]{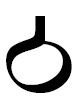}&
\includegraphics[width=0.4 in]{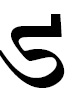}&
\includegraphics[width=0.4 in]{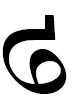}&
\includegraphics[width=0.4 in]{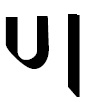}&
\includegraphics[width=0.4 in]{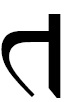}&
\includegraphics[width=0.4 in]{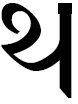}&
\includegraphics[width=0.4 in]{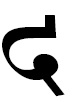}&
\includegraphics[width=0.4 in]{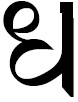}&
\includegraphics[width=0.4 in]{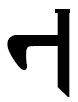}\\
17&18&19&20&21&22&23&24&25&26\\
\includegraphics[width=0.4 in]{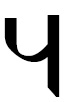}&
\includegraphics[width=0.4 in]{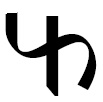}&
\includegraphics[width=0.4 in]{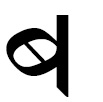}&
\includegraphics[width=0.4 in]{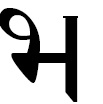}&
\includegraphics[width=0.4 in]{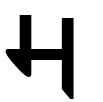}&
\includegraphics[width=0.4 in]{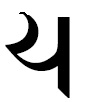}&
\includegraphics[width=0.4 in]{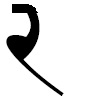}&
\includegraphics[width=0.4 in]{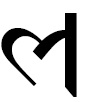}&
\includegraphics[width=0.4 in]{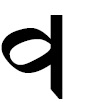}&
\includegraphics[width=0.4 in]{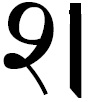}\\
27&28&29&30&31&32&33&34&35&36\\
\includegraphics[width=0.4 in]{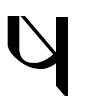}&
\includegraphics[width=0.4 in]{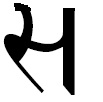}&
\includegraphics[width=0.4 in]{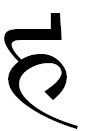}\\
37&38&39&&&&&&&\\
\end{array}$
\caption{Hindi consonants included as part of the recognized classes.}
\end{center}
\end{figure}
Pure consonants or half consonants are the corresponding consonants with their implicit vowel muted. Figure 3 shows subset of Hindi pure consonants considered.
\begin{figure}[ht]
\begin{center}
$\begin{array}{cccccccccc}
\includegraphics[width=0.45 in]{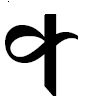}&
\includegraphics[width=0.45 in]{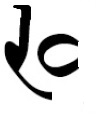}&
\includegraphics[width=0.4 in]{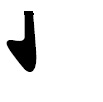}&
\includegraphics[width=0.4 in]{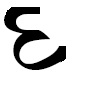}&
\includegraphics[width=0.4 in]{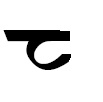}&
\includegraphics[width=0.4 in]{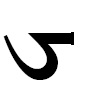}&
\includegraphics[width=0.4 in]{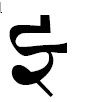}&
\includegraphics[width=0.4 in]{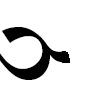}&
\includegraphics[width=0.4 in]{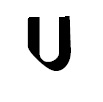}&
\includegraphics[width=0.4 in]{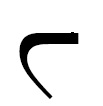}\\
40&41&42&43&44&45&46&47&48&49\\
\includegraphics[width=0.4 in]{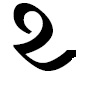}&
\includegraphics[width=0.4 in]{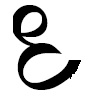}&
\includegraphics[width=0.4 in]{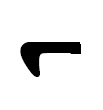}&
\includegraphics[width=0.4 in]{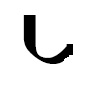}&
\includegraphics[width=0.4 in]{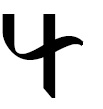}&
\includegraphics[width=0.4 in]{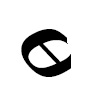}&
\includegraphics[width=0.4 in]{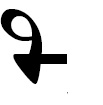}&
\includegraphics[width=0.4 in]{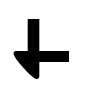}&
\includegraphics[width=0.4 in]{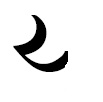}&
\includegraphics[width=0.4 in]{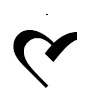}\\
50&51&52&53&54&55&56&57&58&59\\
\includegraphics[width=0.4 in]{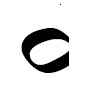}&
\includegraphics[width=0.4 in]{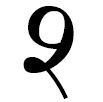}&
\includegraphics[width=0.4 in]{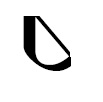}&
\includegraphics[width=0.4 in]{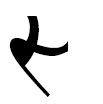}&
\includegraphics[width=0.4 in]{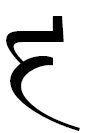}\\
60&61&62&63&64&&&&&\\
\end{array}$
\caption{Hindi half consonants considered for recognition.}
\end{center}
\end{figure}

The nasalization sign indicates nasalization of the character the sign is written over. The vowel omission sign is used to mute the implicit vowel of a consonant. The vowel sign is used to modify the implicit vowel of a consonant. Figure 4(65) shows the nasalization sign, Figure 4(66) shows the vowel omission sign, Figures 4(67)-(72) show vowel signs and Figure 4(73) shows a consonant modified by a vowel sign.
\begin{figure}[ht!]
\begin{center}
$\begin{array}{ccccccccc}
\includegraphics[width=0.45 in]{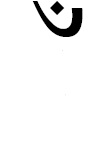}&
\includegraphics[width=0.45 in]{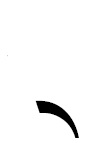}&
\includegraphics[width=0.4 in]{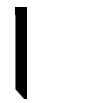}&
\includegraphics[width=0.4 in]{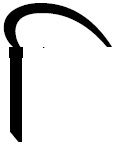}&
\includegraphics[width=0.4 in]{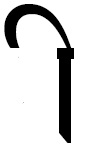}&
\includegraphics[width=0.4 in]{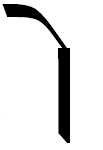}&
\includegraphics[width=0.4 in]{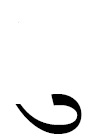}&
\includegraphics[width=0.4 in]{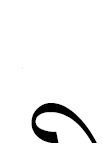}&
\includegraphics[width=0.4 in]{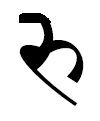}\\
65&66&67&68&69&70&71&72&73\\
\end{array}$
\caption{Some signs and consonant with vowel sign considered for recognition. (65) Nasalization sign. (66) Vowel omission sign. (67)-(72) Vowel signs. (73) Consonant with vowel sign.}
\end{center}
\end{figure}
Conjuncts are cluster of consonants where implicit vowels of all but the last consonant of the cluster are muted. Some conjuncts considered are shown in Figure 5.
\begin{figure}[ht!]
\begin{center}
$\begin{array}{cccccccccc}
\includegraphics[width=0.4 in]{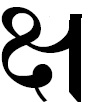}&
\includegraphics[width=0.4 in]{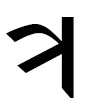}&
\includegraphics[width=0.4 in]{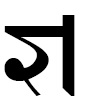}&
\includegraphics[width=0.4 in]{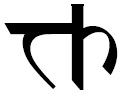}&
\includegraphics[width=0.4 in]{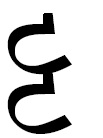}&
\includegraphics[width=0.4 in]{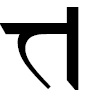}&
\includegraphics[width=0.4 in]{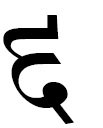}&
\includegraphics[width=0.4 in]{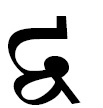}&
\includegraphics[width=0.4 in]{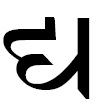}&
\includegraphics[width=0.4 in]{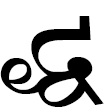}\\
74&75&76&77&78&79&80&81&82&83\\
\includegraphics[width=0.4 in]{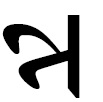}&
\includegraphics[width=0.4 in]{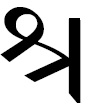}&
\includegraphics[width=0.4 in]{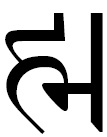}&
\includegraphics[width=0.4 in]{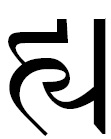}&
\includegraphics[width=0.4 in]{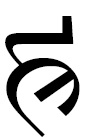}&
\includegraphics[width=0.4 in]{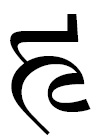}&
\includegraphics[width=0.4 in]{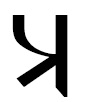}&
\includegraphics[width=0.4 in]{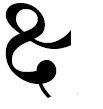}&
\includegraphics[width=0.4 in]{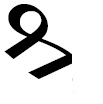}&
\includegraphics[width=0.4 in]{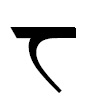}\\
84&85&86&87&88&89&90&91&92&93\\
\end{array}$
\caption{Conjuncts and consonant clusters included as part of the recognized classes. Conjuncts are clusters of consonants, where only the implicit vowel of the last consonant is not muted.(74)-(90) Frequently used conjuncts. Symbols, where the implicit vowel of the last consonant of a consonant cluster is muted, are also used. Symbols 91 to 93 belong to this category.}
\label{conjunc}
\end{center}
\end{figure}
Some strokes with different shapes are shown in Figure 6.  
\begin{figure}[ht]
\begin{center}
$\begin{array}{ccc}
\includegraphics[width=0.45 in]{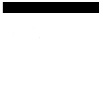}&
\includegraphics[width=0.45 in]{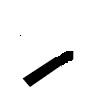}&
\includegraphics[width=0.4 in]{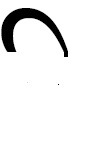}\\
94&95&96\\
\end{array}$
\caption{Other strokes with different shapes considered for recognition.}
\end{center}
\end{figure}
\subsection{Hindi character dataset}
\indent Samples of Hindi online handwritten characters have been collected from Center for Development of Advanced Computation (CDAC) and HP datasets \cite{trohp}. Characters that are not cursive and have been written correctly are collected. Figures 7(a) and 7(b) show samples of Hindi online handwritten characters obtained from the constructed dataset. These characters are produced as a sequence of strokes in a particular order. The strokes are produced in a particular direction. The header lines over all the characters have been removed because they do not contribute to the structure of the characters.\\
\indent The total number of samples of characters in the dataset is 15653. The dataset is divided to form training and testing datasets. The training and testing datasets consist of 12832 and 2821 samples, respectively. These datasets consist of samples from 96 different characters with an average of 133 and 29 samples per character class in the training and testing datasets, respectively.
\begin{figure}[ht]
\begin{center}
$\begin{array}{cc}
\includegraphics[width=2.5 in]{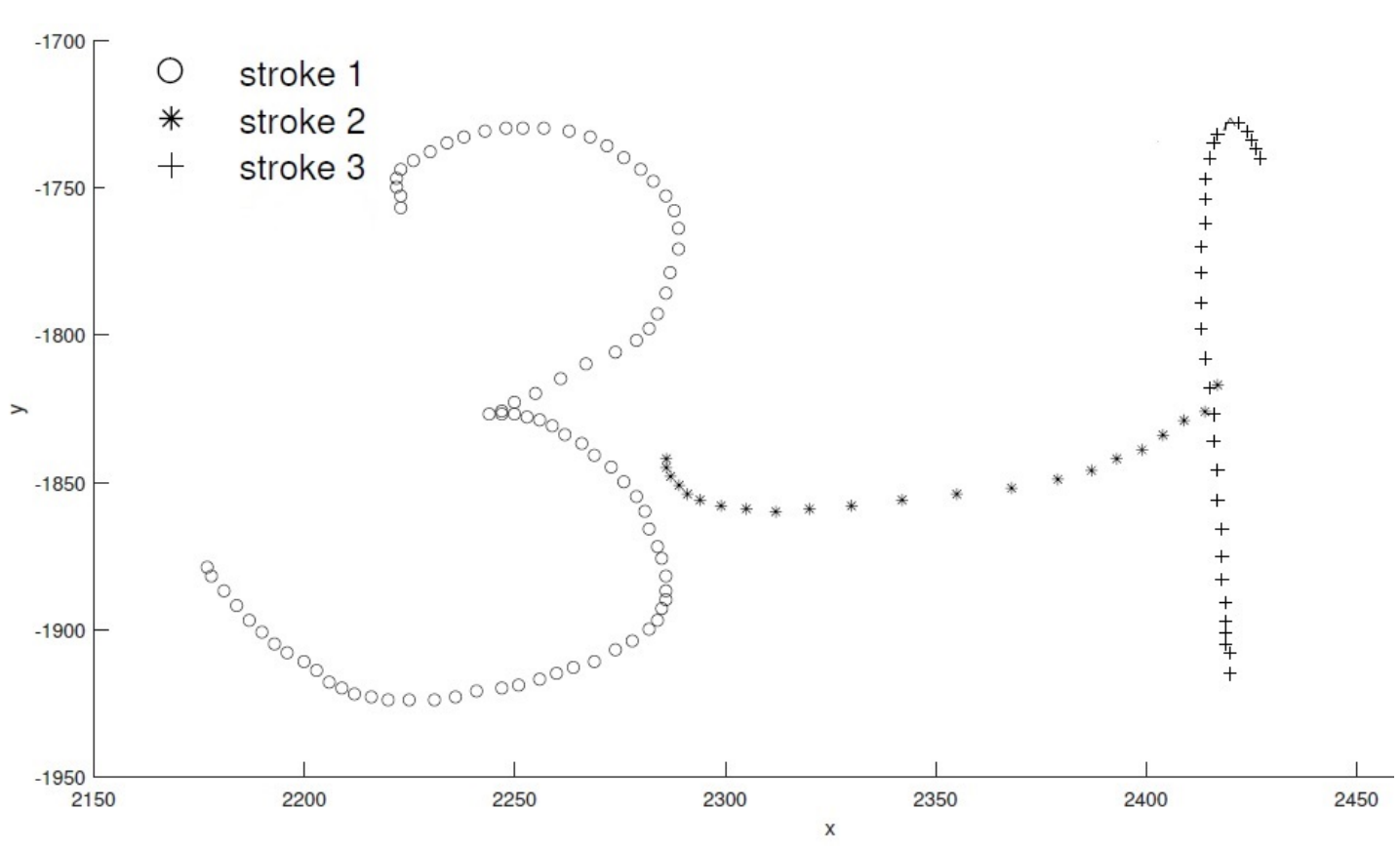}&
\includegraphics[width=2 in]{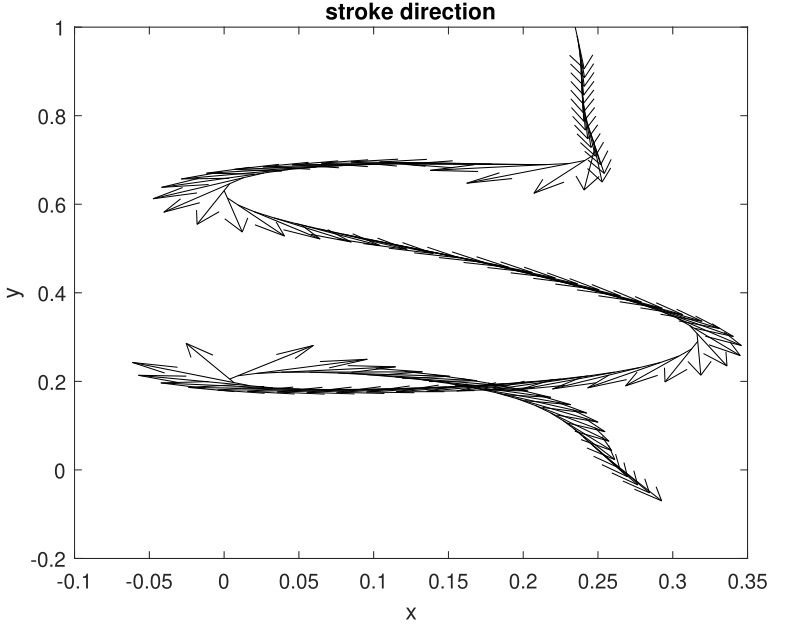}\\

(a)&(b)\\
\end{array}$
\end{center}
\caption{(a) Sample of Hindi online handwritten character produced using three strokes in the order stroke1, stroke2 and stroke3. (b) Sample of Hindi online handwritten character produced in the direction shown by the arrows.}
\end{figure}
\subsection{Preprocessing of samples in the dataset}
\indent The samples of online handwritten characters in the constructed datasets have got variations because they have been produced by different individuals at different times. Variations among samples of characters can be classified as external variations and internal variations. External variations are variations that can be removed by preprocessing or feature design. Internal variations are caused by deviation of handwritten character structure from the ideal character structure and are very difficult to remove. It is important to preprocess the  characters so that feature extraction from characters can be done in a uniform fashion.\\
\indent Some of the external variations removed from handwritten characters in the datasets are as follows. Repeated points are consecutive points in a character that have the same co-ordinate values and do not contribute to the structure of the character and so are detected and removed. Co-ordinate values of points in characters in both the $x$ and $y$ directions are mapped to the range $[0\,\,1]$ by linear transformation. This removes variations in location and size of characters. Distance between all the consecutive points in characters are made equal to a constant $\Delta$ thus removing variations in speed at which handwritten characters have been produced. Character stroke trace roughness is removed by linear filtering the sequences of x-co-ordinates and y-co-ordinates of points in the characters.\\

\subsection{Preprocessed character dataset}
\indent After preprocessing, an online handwritten character is represented as a sequence of strokes $C=(S_1,\dots,S_{N_C})$, where a stroke is represented as\[S_i=[{p_1^{S_i}}^T;\dots ;{p_n^{S_i}}^T;\dots ;{p_{N_{S_i}}^{S_i}}^T],\quad S_i\in[0\,\, 1]^{N_{S_i}\times2},\quad 1\le i\le N_C.\] Point $p_n^{S_i}=[x_n\,\, y_n]^T,\,\,\, p_n^{S_i}\in[0 \,\,1]^{2\times1}$, is the $n^{th}$ point in the stroke $S_i$ and $x_n$ and $y_n$ are, respectively, the $x$-co-ordinate and $y$-co-ordinate of the $n^{th}$ point in the stroke $S_i$. $N_C$ is the number of strokes in the character $C$. Figure 8 shows samples of online handwritten characters before and after preprocessing.  
\begin{figure}[ht]
\begin{center}
$\begin{array}{cc}
\includegraphics[width=2.5 in]{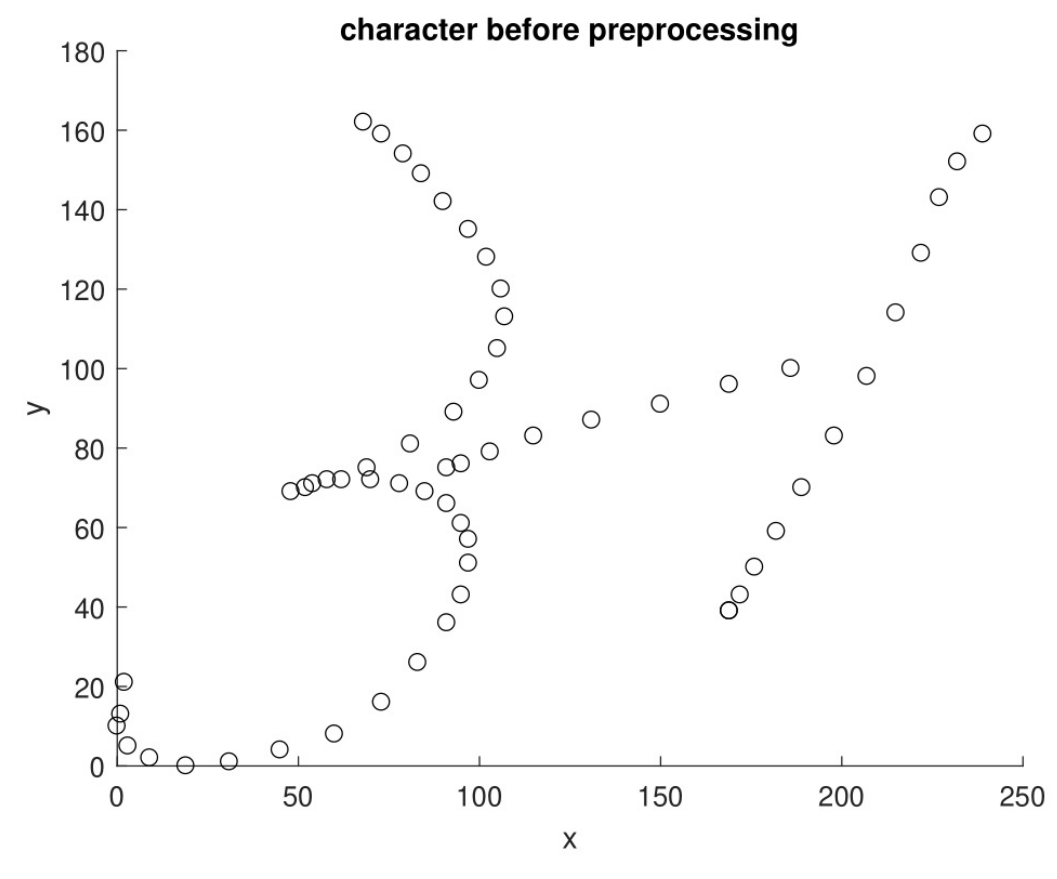}&
\includegraphics[width=2.5 in]{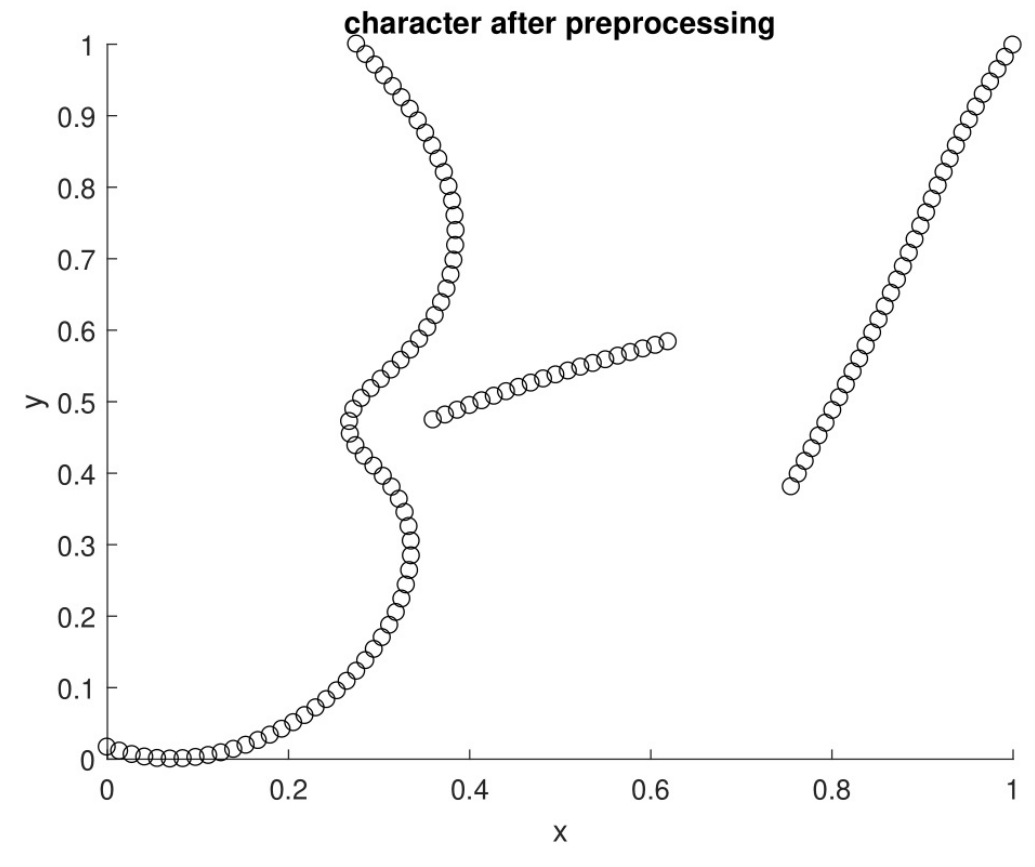}\\
(a)&(b)\\
\includegraphics[width=2.5 in]{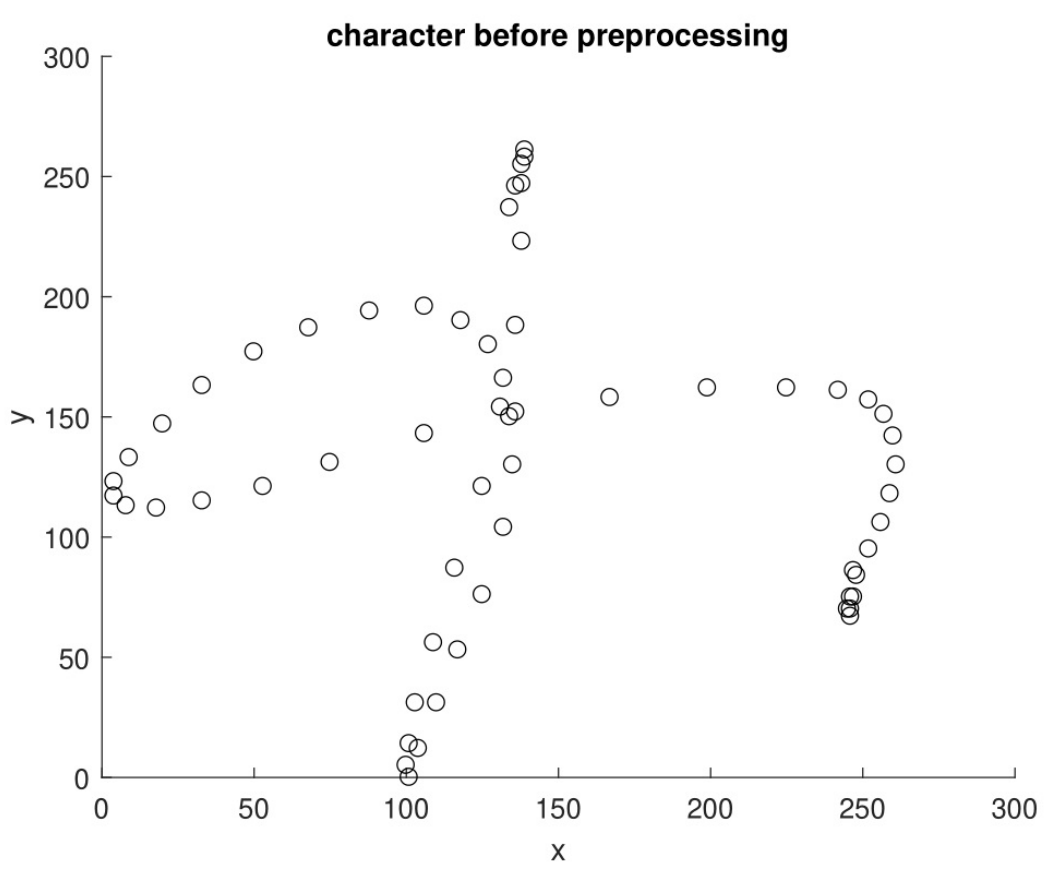}&
\includegraphics[width=2.5 in]{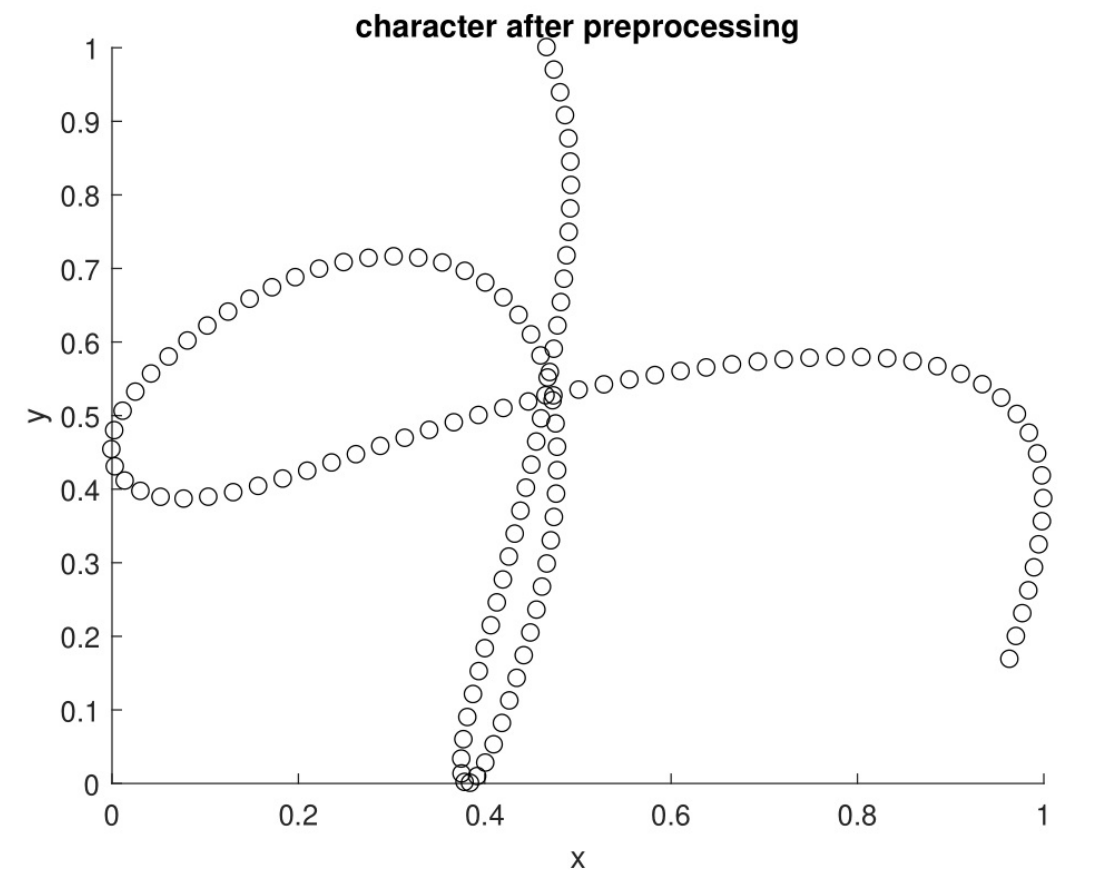}\\
(c)&(d)\\
\end{array}$
\caption{Characters before and after preprocessing. (a) and (c) Two characters before preprocessing. (b) and (d) The same characters, after application of preprocessing steps like removal of repeated points, variations in location and size, variation in distance between consecutive points, and roughness of trace of the characters.}
\end{center}
\end{figure}
Distance $\Delta$ between consecutive points in these characters is chosen to be 0.02. Number of points for the representation of characters is 128. When the vectors of x- and y-co-ordinates of these points are concatenated, the vector of these spatio-temporal features has a size of 256. The span of a character along the x- and y-co-ordinates are also considered as features. The size of the spatio-temporal feature vector after considering the span features becomes 258. These span features are considered in the construction of different feature vectors by Sharma \cite{hpod}, where the smallest size of the extracted feature vector is 258 and the largest is 786. It is suggested by Duda et al. \cite{troam} to have number of samples per character class more than the size of the feature vectors in that character class. In this work the average number of samples per character class in the training dataset is 133 which is smaller than the size of the feature vectors to be used for training the classifier. Therefore the classifiers considered for performance comparison with that of the developed classifier are the ones that can be trained on a small training dataset and can produce good classification performance on the corresponding testing dataset.             

\section{Classifier design}  
\indent Samples of on-line handwritten characters of same class are different because of the internal and external variations and these variations are called the intra-class variations. Samples of on-line handwritten characters of different classes are different because of the structural variations and these variations are called the inter-class variations. External variations can be removed but internal variations are difficult to remove. Internal variations can be captured using statistical models. Classifier models considered in this study are all statistical models in the sense that their parameters are determined from samples of online handwritten characters. These classifiers can correctly classify character samples to their correct class if intra-class variations are less than the inter-class variations.\\
\indent A classifier $H_{w}$ defined in terms of parameters $w$ is a function from feature space to class label space, $H_w:\mathcal{X}\rightarrow \mathcal{Y}$. The classifier parameters are estimated from the samples of handwritten characters. A classifier is said to be determined when the parameters have been estimated. Once the classifier is determined it is used to classify samples of handwritten characters. A classifier is said to have a good classification performance or good generalization capability if it correctly classifies samples of handwritten characters that its parameters have not been estimated from.\\
\indent Let $D^{tr}=\{d^{tr}_1,\dots,d^{tr}_{N_{ct}}\}$ be the training set of sets of different character samples where $N_{ct}$ is the number of different character classes or character categories. Let $d^{tr}_k=\{X^{tr}_{k,1},\dots,X^{tr}_{k,|d^{tr}_k|}\},\,\, 1\le k\le N_{ct}$, be the training set of feature vectors of the $k^{th}$ character class where $|d^{tr}_k|$ is the number of samples in the training set of the $k^{th}$ character class. If all the training samples of all the classes are collected along with their class labels then the training set is represented as 
\[\{X^{tr}_m,Y^{tr}_m\}_{m=1}^{N^{tr}},\,\,\,X^{tr}_m\in\mathcal{X},\,\,\,Y^{tr}_m\in\mathcal{Y},\,\,\,N^{tr}=\sum_{k=1}^{N_{ct}}|d^{tr}_k|.\] Here class label of the $m^{th}$ sample $X^{tr}_m$ is $Y^{tr}_m$ where $Y^{tr}_m$ has $1$-of-$N_{ct}$ representation so that \[Y^{tr}_m\in\{0,1\}^{N_{ct}},\quad Y^{tr}_m(k)\in\{0,1\},\quad \sum_{k=1}^{N_{ct}}Y^{tr}_m(k)=1.\]According to this representation the feature vector $X^{tr}_m$ belongs to the $k^{th}$ class if\[Y^{tr}_m(k)=1,\quad Y^{tr}_m(k')=0,\quad k\ne k',\quad \forall k,k'\in\{1,\dots,N_{ct}\},\quad 1\le m\le N^{tr}.\]Similarly, let $D^{ts}=\{d^{ts}_1,\dots,d^{ts}_{N_{ct}}\}$ be the test set of sets of different character samples. Let $d^{ts}_k=\{X^{ts}_{k,1},\dots,X^{ts}_{k,|d^{ts}_k|}\},\,\,\,1\le k\le N_{ct}$, be the test set of feature vectors of the $k^{th}$ character class where $|d^{ts}_k|$ is the number of samples in the test set of $k^{th}$ character class. If all the test samples of all the classes are collected along with their class labels then the test set is represented as \[\{X^{ts}_m,Y^{ts}_m\}_{m=1}^{N^{ts}},\,\,\,X^{ts}_m\in\mathcal{X},\,\,\,Y^{ts}_m\in\mathcal{Y},\,\,\,N^{ts}=\sum_{k=1}^{N_{ct}}|d^{ts}_k|.\] Here class label of the $m^{th}$ sample $X^{ts}_m$ is $Y^{ts}_m$ where $Y^{ts}_m$ has $1$-of-$N_{ct}$ representation so that \[Y^{ts}_m\in\{0,1\}^{N_{ct}},\quad Y^{ts}_m(k)\in\{0,1\},\quad\sum_{k=1}^{N_{ct}}Y^{ts}_m(k)=1.\] According to this representation the feature vector $X^{ts}_m$ belongs to the $k^{th}$ class if\[Y^{ts}_m(k)=1,\quad Y^{ts}_m(k')=0,\quad k\ne k',\quad\forall k,k'\in\{1,\dots,N_{ct}\},\quad 1\le m\le N^{ts}.\]
A classifier $H_{w}$ is estimated from training samples $\{X^{tr}_m,Y^{tr}_m\}_{m=1}^{N^{tr}}$ by minimizing some loss function. Loss function is a non-negative real valued function of classifier output on feature vector and the corresponding class label, $L:\mathcal{Y}\times\mathcal{Y}\rightarrow \mathcal{R}^+$. It would be nice to estimate parameters of $H_w$ by minimizing the expected loss or risk $E\left(L(H_w(X),Y)\right)$ but the joint distribution of $X$ and $Y$ is not available. The parameters of $H_w$ are estimated instead by minimizing empirical risk. Empirical risk is given by $R_{N^{tr}}(H_w)=(N_{tr})^{-1}\sum_{m=1}^{N^{tr}}L(H_w(X^{tr}_m),Y^{tr}_m)$. Classifier $H_w$ is said to have correctly classified a feature vector $X^{ts}_m$ if $H_w(X^{ts}_m)=Y^{ts}_m$. The accuracy of the classifier is the ratio of the total number of correct classifications to the total number of classifications and is given by
\[H^a_w=\frac{\sum_{m=1}^{N^{ts}}I(H_w(X^{ts}_m)=Y^{ts}_m)}{N^{ts}},\,\,\,I(x)=1\,\,\,\mbox{if}\,\,\,x\,\,\,\mbox{is true},\,\,\, I(x)=0\,\,\,\mbox{if}\,\,\,x\,\,\,\mbox{is false}.\] 
The sections below describe some important existing classifiers and the classifier developed in this thesis.

\subsection{Second order statistics (SOS) classifier}
In this model, character samples from each class are assumed to be generated from a Gaussian distribution with parameters $\mu_k,\,\,\,\Sigma_k,\,\,\,1\le k\le N_{ct}$ as given in Duda et al.  \cite{troam}. Here, $\mu_k$ and $\Sigma_k$ are, respectively, the mean and co-variance matrix of the $k^{th}$ class distribution. These parameters are estimated from the features extracted from the character samples as
\[\mu_k=\frac{\sum_{m=1}^{|d^{tr}_k|}X^{tr}_{k,m}}{|d^{tr}_k|},\quad X^{tr}_{k,m}\in\mathcal{R}^{N_{ftr}\times 1},\quad 1\le k\le N_{ct},\]
\[\Sigma_k=\frac{\sum_{m=1}^{|d^{tr}_k|}(X^{tr}_{k,m}-\mu_k)(X^{tr}_{k,m}-\mu_k)^T}{|d^{tr}_k|},\quad 1\le k\le N_{ct}.\]
$N_{ftr}$ is the size of the feature vector and $w_k=(\mu_k,\Sigma_k),\,\,\,1\le k\le N_{ct},\,\,\,w=(w_1,\dots,w_{N_{ct}})$ are the parameters of the model.
Likelihood of a test feature vector $X^{ts}_m$ being generated by the $k^{th}$ class distribution is
\[H^{SOS}_{w_k}(X^{ts}_m)=\frac{1}{(2\pi)^{\frac{N_{ftr}}{2}}|\Sigma_k|^{\frac{1}{2}}}\exp\left(-\frac{1}{2}{(X^{ts}_{m}-\mu_k)^T\Sigma_k^{-1}(X^{ts}_{m}-\mu_k)}\right),\,\,\,1\le k\le N_{ct}.\]
The decision of SOS classifier on the class label of the feature vector $X^{ts}_{m}\in\mathcal{R}^{N_{ftr}\times 1},\,\,\,1\le m\le N_{ts}$, is
\[H^{SOS}_{w,m}(k)=\begin{cases}1,&\mbox{if}\,\,\,k=\underset{k'}{arg\,max}\,\, H^{SOS}_{w_{k'}}(X^{ts}_m)\\0,&\mbox{otherwise}\end{cases},\,\,\,\mbox{for}\,\,\,1\le k,k'\le N_{ct},\,\,\,H^{SOS}_{w,m}\in\{0,1\}^{N_{ct}}.\]
The accuracy of the classifier is\,\,\,
\[H^{a,SOS}_w=\frac{\sum_{m=1}^{N^{ts}}I(H^{SOS}_{w,m}=Y^{ts}_m)}{N^{ts}}.\]

\subsection{Sub-space (SS) classifier}
In this model, character samples from each class are assumed to have a representation in sub-space having small dimension compared to the high dimensional vector space in which the samples are observed as given in Duda et al. \cite{troam}. To find the sub-space representation for each character class, mean $\mu_k$ and co-covariance matrix $\Sigma_k$ are computed from the features extracted from the character samples from each character class.
\[\mu_k=\frac{\sum_{m=1}^{|d^{tr}_k|}X^{tr}_{k,m}}{|d^{tr}_k|},\quad X^{tr}_{k,m}\in\mathcal{R}^{N_{ftr}\times 1},\quad 1\le k\le N_{ct},\]
\[\Sigma_k=\frac{\sum_{m=1}^{|d^{tr}_k|}(X^{tr}_{k,m}-\mu_k)(X^{tr}_{k,m}-\mu_k)^T}{|d^{tr}_k|},\quad 1\le k\le N_{ct}.\]
The vectors required for the construction of sub-space for each character class are obtained by solving the eigen value problem $\Sigma_k\,\,\phi^k=\lambda^k\,\phi^k$

to get the eigen values  $\lambda^k_{1},\dots,\lambda^k_{N_{ftr}}$ and the corresponding eigen vectors $\phi^k_{1},\dots,\phi^k_{N_{ftr}},\,\,\,1\le k\le N_{ct}$.  The sub-space representation for the $k^{th}$ class is constructed by considering the $n_{ef}$ largest eigen values $\lambda^k_{\pi(1)},\dots,\lambda^k_{\pi(n_{ef})}$ and the corresponding eigen vectors $\phi^k_{\pi(1)},\dots,\phi^k_{\pi(n_{ef})}$ for $1\le k\le N_{ct}$.
These eigen vectors are collected to form classifier with parameters $w=(\{\phi^k_{\pi(1)},\dots,\phi^k_{\pi(n_{ef})}\}_{k=1}^{N_{ct}},n_{ef})$. Projection of a test feature vector onto the constructed sub-space for $k^{th}$ class is
\[X^{prj,ts}_{k,m}=\sum_{n=1}^{n_{ef}}\left({X^{\mu,ts}_m}^T\phi^k_{\pi(n)}\right)\phi^k_{\pi(n)},\,\,\,X^{\mu,ts}_m=\left({X^{ts}_m}-\mu_k\right),\,\,1\le k\le N_{ct},\,\,\,1\le m\le N^{ts}.\]
The decision of SS classifier on the class label of the feature vector $X^{ts}_{m}\in\mathcal{R}^{N_{ftr}\times 1},\,\,\,1\le m\le N_{ts}$, is
\[H^{SS}_{w,m}(k)=\begin{cases}1,&\mbox{if}\,\,\,k=\underset{k'}{arg\,min}\,\, ||X^{\mu,ts}_m-X^{prj,ts}_{k',m}||_2\\0,&\mbox{otherwise}\end{cases},\,\,\,\mbox{for}\,\,\,1\le k,k'\le N_{ct},\,\,\,H^{SS}_{w,m}\in\{0,1\}^{N_{ct}}.\]
\[\mbox{The accuracy of classifier is}\,\,\,
H^{a,SS}_w=\frac{\sum_{m=1}^{N^{ts}}I(H^{SS}_{w,m}=Y^{ts}_m)}{N^{ts}}.\]

\subsection{Fisher discriminant (FD) classifier}
The character samples are modeled by their projections onto a sub-space where they are maximally separable as given in Duda et al. \cite{troam}. This sub-space is common to all the character classes and involves training set character samples from all the classes for its determination.   
\[\mbox{Mean of the feature vectors of}\,\,\, k^{th}\,\,\,\mbox{class is}\,\,\, \mu_k=\frac{\sum_{m=1}^{|d^{tr}_k|}X^{tr}_{k,m}}{|d^{tr}_k|},\quad 1\le k\le N_{ct}.\]
\[\mbox{Mean of all the feature vectors of all the classes is}\,\,\,\mu_T=\frac{\sum_{k=1}^{N_{ct}}|d^{tr}_k|\mu_k}{\sum_{k=1}^{N_{ct}}|d^{tr}_k|}.\]
\[\mbox{Scatter matrix of the}\,\,\, k^{th}\,\ \mbox{class is}\,\, C^{wi}_k=\sum_{m=1}^{|d^{tr}_k|}(X^{tr}_{k,m}-\mu_k)(X^{tr}_{k,m}-\mu_k)^T,\quad 1\le k\le N_{ct}.\]
\[\mbox{The within-class scatter matrix is}\,\,\, C^{wi}=\sum_{k=1}^{N_{ct}}C^{wi}_k.\]
\[\mbox{The between-class scatter matrix is}\,\,\, C^{bt}=\sum_{k=1}^{N_{ct}}\,|d^{tr}_k|(\mu_k-\mu_T)(\mu_k-\mu_T)^T.\]     
Let $w_k,\,\,\,1\le k\le N_{ct}-1,\,\,\,N_{ct}\le N_{ftr}$, be the vectors spanning the required sub-space. Let $W$ be the matrix obtained by collecting the vectors $w_k,\,\,\,1\le k\le N_{ct}-1$, as its column vectors. Then the scatter matrix for the projected $k^{th}$ class is
\[\tilde{C}^{wi}_k=\sum_{m=1}^{|d^{tr}_k|}W^T(X^{tr}_{k,m}-\mu_k)(X^{tr}_{k,m}-\mu_k)^TW,\quad 1\le k\le N_{ct}.\]
\[\mbox{The projected within-class scatter matrix is}\,\,\, \tilde{C}^{wi}=\sum_{k=1}^{N_{ct}}W^TC^{wi}_kW=W^TC^{wi}W.\]
The projected between-class scatter matrix is\[\tilde{C}^{bt}=\sum_{k=1}^{N_{ct}}\,|d^{tr}_k|W^T(\mu_k-\mu_T)(\mu_k-\mu_T)^TW=W^TC^{bt}W.\] 
The required sub-space vectors $W$ can be found by maximizing the ratio
\[J(W)=\frac{\det(\tilde{C}^{bt})}{\det(\tilde{C}^{wi})}=\frac{\det(W^TC^{bt}W)}{\det(W^TC^{wi}W)}.\]
The solution to the above problem is obtained as a solution to generalized eigen value problem $C^{bt}\phi=\lambda\,\, C^{wi}\phi$. This problem is solved to get $N_{ct}-1$ eigen vectors $\phi_{1},\dots,\phi_{N_{ct}-1}$ corresponding to the $N_{ct}-1$ largest eigen values $\lambda_{1},\dots,\lambda_{N_{ct}-1}$. These vectors are used to construct the Fisher matrix $W$ after being normalized using Gram-Schmidt procedure. Following this construction, all the character feature vectors are projected on to the space spanned by $W$ vectors. The projected feature vectors for the characters from training set are obtained as\[X^{tr,F}_{k,m}=W^T\,X^{tr}_{k,m},\,\,\,X^{tr,F}_{k,m}\in\mathcal{R}^{(N_{ct}-1)\times 1},\,\,\,1\le m\le |d^{tr}_k|,\,\,\,1\le k\le N_{ct}.\]
The projected feature vectors for the characters from test set are obtained as\[X^{ts,F}_{k,m}=W^T\,X^{ts}_{k,m},\,\,\,X^{ts,F}_{k,m}\in\mathcal{R}^{(N_{ct}-1)\times 1},\,\,\,1\le m\le |d^{ts}_k|,\,\,\,1\le k\le N_{ct}.\]
These feature vectors are assumed to be generated from Gaussian distribution with parameters $\mu^F_k,\,\,\Sigma^F_K,\,\,1\le k\le N_{ct}$, which are estimated as
\[\mu^F_k=\frac{\sum_{m=1}^{|d^{tr}_k|}X^{tr,F}_{k,m}}{|d^{tr}_k|},\quad 1\le k\le N_{ct},\]
\[\Sigma^F_k=\frac{\sum_{m=1}^{|d^{tr}_k|}(X^{tr,F}_{k,m}-\mu^F_k)(X^{tr,F}_{k,m}-\mu^F_k)^T}{|d^{tr}_k|},\quad 1\le k\le N_{ct},\]
\[w_k=(\mu^F_k,\Sigma^F_k),\,\,\,1\le k\le N_{ct},\,\,\,w=(w_1,\dots,w_{N_{ct}}).\]
Likelihood of a test feature vector being generated from $k^{th}$ class distribution is
\[H^{FD}_{w_k}(X^{ts,F}_m)=\frac{1}{(2\pi)^{\frac{(N_{ct}-1)}{2}}|\Sigma^F_k|^{\frac{1}{2}}}\exp\left(-\frac{1}{2}{(X^{ts,F}_{m}-\mu^F_k)^T({\Sigma^F_k})^{-1}(X^{ts,F}_{m}-\mu^F_k)}\right),\,\,\,1\le k\le N_{ct}.\]
The decision of FD classifier on the class label of the feature vector $X^{ts,F}_m,\,\,\,1\le m\le N_{ts}$, is
\[H^{FD}_{w,m}(k)=\begin{cases}1,&\mbox{if}\,\,\,k=\underset{k'}{arg\,max}\,\, H^{FD}_{w_{k'}}(X^{ts,F}_m)\\0&\mbox{otherwise}\end{cases},\,\,\,\mbox{for}\,\,\,1\le k,k'\le N_{ct},\,\,\,H^{FD}_{w,m}\in\{0,1\}^{N_{ct}}.\]
\[\mbox{The accuracy of classifier is}\,\,\,
H^{a,FD}_w=\frac{\sum_{m=1}^{N^{ts}}I(H^{FD}_{w,m}=Y^{ts}_m)}{N^{ts}}.\]

\subsection{Feedforward neural network (FNN) classifier}
The feedforward neural network as described in Haykin \cite{troak}, to be used for classification of character samples, consists of three layers: input layer, hidden layer and output layer. There are $N_{ftr}$ number of inputs in the input layer corresponding to the size of feature vectors. Feature vectors in the training and testing sets are used for training and testing the classifier. There are $N_{hid}$ number of neurons or computational units in the hidden layer each connected to all the inputs in the input layer. There are $N_{ct}$ number of computational units in the output layer each connected to all the computational units in the hidden layer. The classifier has to classify character feature vectors into one of the $N_{ct}$ character classes so the feedforward network has $N_{ct}$ number of computational units at its output. Overall the feedforward neural network is a collection of interconnected computational units that acts as nonlinear functional mapping from input to output.\\
\indent The functional mapping is learned from training set of handwritten character samples by propagating the corresponding feature vectors forward through the network and propagating error backwards through the network. Each computational unit in the hidden layer computes its output by taking the inner product of the input feature vector $X^{tr}_m,\,\,\,1\le m\le N^{tr}$, and the weights of the connections between inputs and that computational unit followed by transformation of the inner product using differentiable nonlinear function and is given by
\[w^h_j=\sum_{n=1}^{N_{ftr}}w_{j,n}\,X^{tr}_m(n)+w_{j0},\quad w^{h,o}_j=\sigma_h(w^h_j),\quad 1\le j\le N_{hid}.\]  
Each computational unit in the output layer computes its output by taking the inner product of the outputs from the hidden units $w^{h,o}_j,\,\,\,1\le j\le N_{hid},$ and the weights of the connections between the hidden units and that output unit followed by transformation of the inner product by differentiable nonlinear function and is given by
\[w_k^o=\sum_{j=1}^{N_{hid}}w_{k,j}\,w^{h,o}_j+w_{k0},\quad Y_k^o=\sigma_o(w^o_k),\quad 1\le k\le N_{ct}.\]   
\[\mbox{Here}\,\,\, w_{j0},\,\,\,1\le j\le N_{hid},\,\,\, w_{k0},\,\,\,1\le k\le N_{ct},\,\,\,\mbox{are the bias terms}.\] The outputs $Y^o_k,\,\,\,1\le k\le N_{ct}$, are functions of all the parameters $w=(w_{j,n},\,\,\,w_{j0},\,\,\,1\le j\le N_{hid},\,\,\,1\le n \le N_{ftr},\,\,\,w_{k,j},\,\,\,w_{k0},\,\,\,1\le j\le N_{hid},\,\,\,1\le k \le N_{ct})$ of the network and so can be written as $Y^o_k(X^{tr}_m,w)$. The parameters $w$ are estimated by minimizing the error function $E(w)$ by using back-propagation algorithm.
\[E(w)=\sum_{m=1}^{N^{tr}}E(X^{tr}_m,w),\,\, \mbox{where}\,\,  E(X^{tr}_m,w)=\frac{1}{2}\sum_{k=1}^{N_{ct}}\left(Y^o_k(X^{tr}_m,w)-Y^{tr}_m(k)\right)^2.\]
Once the parameters of the network have been estimated, FNN can be used for classification of character samples.
\[\mbox{The decision of the classifier on the class label of the feature vector}\,\,\, X^{ts}_{m}\in\mathcal{R}^{N_{ftr}\times 1},\,\,\,1\le m\le N_{ts},\,\,\,\mbox{is}\]
\[H^{FNN}_{w,m}(k)=\begin{cases}1,&\mbox{if}\,\,\,k=\underset{k'}{arg\,max}\,\, Y^o_{k'}(X^{ts}_m,w),\\0,&\mbox{otherwise}\end{cases},\,\,\,\mbox{for}\,\,\,1\le k,k'\le N_{ct},\,\,\,H^{FNN}_{w,m}\in\{0,1\}^{N_{ct}}.\]
\[\mbox{The accuracy of classifier is}\,\,\,
H^{a,FNN}_w=\frac{\sum_{m=1}^{N^{ts}}I(H^{FNN}_{w,m}=Y^{ts}_m)}{N^{ts}}.\]

\subsection{Support vector machines (SVM) classifier}
\indent SVM classifier finds decision boundary or separating hyperplane between samples of two different classes such that the margin or the minimum distance between the hyperplane and the nearest samples is maximized as described in Cortes et al. \cite{clsfff}. So SVM is also called a maximum margin classifier. Let $(w_1,\dots,w_{N_{ftr}},w_0),\,\,\,w=(w_1,\dots,w_{N_{ftr}})$, be the parameters defining a hyperplane. Let $d^{tr}_k=\{X^{tr}_{k,1},\dots,X^{tr}_{k,|d^{tr}_k|}\}$, and $d^{ts}_k=\{X^{ts}_{k,1},\dots,X^{ts}_{k,|d^{ts}_k|}\},\,\,1\le k\le 2,$ respectively, be the sets of training and testing samples from two character classes. If all the samples of the training set are collected along with their class labels then training set is represented as $\{X^{tr}_m,y^{tr}_m\}_{m=1}^{N'^{tr}},\,\,\,N'^{tr}=|d^{tr}_1|+|d^{tr}_2|$, where $y^{tr}_m=1$ for class one samples and $y^{tr}_m=-1$ for class two samples. Similarly, if all the samples of the testing set are collected along with their class labels then testing set is represented as $\{X^{ts}_m,y^{ts}_m\}_{m=1}^{N'^{ts}},\,\,\,N'^{ts}=|d^{ts}_1|+|d^{ts}_2|$, where $y^{ts}_m=1$ for class one samples and $y^{ts}_m=-1$ for class two samples. If the samples from two classes are linearly separable then the hyperplane defined by $(w,w_0)$ will be able to separate all the $m,\,\,\,1\le m\le N'^{tr}$, samples as
\[w^T\,X^{tr}_m+w_0>0,\quad \mbox{for}\quad y^{tr}_m=1,\] 
\[w^T\,X^{tr}_m+w_0<0,\quad \mbox{for}\quad y^{tr}_m=-1.\]
The above two classification conditions can be equivalently written as
\[y^{tr}_m\left(w^T\,X^{tr}_m+w_0\right)>0.\]  
Let $y^{p,tr}_m(X^{tr}_m)=w^T\,X^{tr}_m+w_0$, then the above classification conditions can also be written as $y^{tr}_m\,y^{p,tr}_m(X^{tr}_m)>0$.
The perpendicular distance of a sample or a point to the hyperplane $(w,w_0)$ is given by $y^{tr}_m\left(w^T\,X^{tr}_m+w_0\right)\,(||w||_2)^{-1}$. Since scaling of $(w,w_0)$ does not affect the perpendicular distance of any point from the hyperplane, scaling of $(w,w_0)$ is done so that the sample which is at a minimum distance from the decision hyperplane satisfies
\[y^{tr}_m\left(w^T\,X^{tr}_m+w_0\right)=1.\]
Then this decision hyperplane $(w ,w_0)$ will be such that all the samples will satisfy the classification conditions
\[y^{tr}_m\left(w^T\,X^{tr}_m+w_0\right)\ge1,\quad 1\le m\le N'^{tr}.\]
Then the margin is $||w||_2^{-1}$ and SVM is a maximum margin classifier so the separating hyperplane is obtained by maximizing $||w||_2^{-1}$. This is equivalent to minimizing $||w||_2^2$. There is possibility of some samples falling on the wrong side of the separating hyperplane. To account for this possibility, slack variables $\xi_m\ge0,\,\,\,1\le m\le N'^{tr}$, are introduced and then classification conditions are modified as      
\[y^{tr}_m\left(w^T\,X^{tr}_m+w_0\right)\ge1-\xi_m,\quad 1\le m\le N'^{tr}.\]
Now the separating hyperplane is determined by minimizing $||w||_2^2$ by penalizing samples that lie on the wrong side of the separating hyperplane and therefore the objective function to be minimized is
\[\frac{1}{2}w^Tw+\beta\sum_{m=1}^{N'^{tr}}\xi_m,\quad y^{tr}_m\left(w^T\,X^{tr}_m+w_0\right)\ge1-\xi_m,\quad\beta>0,\quad \xi_m\ge 0,\quad 1\le m\le N'^{tr},\]
where $\beta$ is the regularization parameter. This constrained optimization problem is solved by construction of Lagrangian function
\[J(w,w_0,\xi,\varphi)=\frac{1}{2}w^Tw+\beta\sum_{m=1}^{N'^{tr}}\xi_m+\sum_{m=1}^{N'^{tr}}\varphi_m(1-\xi_m-y^{tr}_m\,y^{p,tr}_m(X^{tr}_m))-\sum_{m=1}^{N'^{tr}}\nu_m\xi_m,\]
\[\mbox{where}\,\,\,\phi_m\ge 0,\quad \nu_m\ge 0,\quad 1\le m\le N'^{tr},\,\,\,\mbox{are Lagrange multipliers.}\]
By applying the Karush Kuhn Tucker (KKT) conditions to the above function, the dual Lagrangian is obtained as
\[L^d(\varphi)=\sum_{m=1}^{N'^{tr}}\varphi_m-\frac{1}{2}\sum_{m=1}^{N'^{tr}}\sum_{m'=1}^{N'^{tr}}\varphi_m\,\,\varphi_{m'}\,\,y^{tr}_m\,\,y^{tr}_{m'}\,\,{X^{tr}_m}^TX^{tr}_{m'}.\]
The function $L^d({\varphi})$ is subject to the constraints
\[0\le \varphi_m\le \beta,\quad \sum_{m=1}^{N'^{tr}}\varphi_m\,y^{tr}_m=0,\quad 1\le m\le N'^{tr}.\] 
The feature vectors can be implicitly nonlinearly mapped to a high dimensional feature space using kernel trick so that the inner product ${X^{tr}_m}^TX^{tr}_{m'}$ can be replaced by $k^r(X^{tr}_m,X^{tr}_{m'})$ with the requirement that the matrix $K^r(m,m')=k^r(X^{tr}_m,X^{tr}_{m'}),\,\,\,1\le m,m'\le N'^{tr},$ is positive definite. Then the dual Lagrangian function can be written as
\[L^d(\varphi)=\sum_{m=1}^{N'^{tr}}\varphi_m-\frac{1}{2}\sum_{m=1}^{N'^{tr}}\sum_{m'=1}^{N'^{tr}}\varphi_m\,\,\varphi_{m'}\,\,y^{tr}_m\,\,y^{tr}_{m'}\,\,k^r({X^{tr}_m},X^{tr}_{m'}).\]
with the constraints
\[0\le \varphi_m\le \beta,\quad \sum_{m=1}^{N'^{tr}}\varphi_m\,y^{tr}_m=0,\quad 1\le m\le N'^{tr}.\] 
The parameters $\varphi=(\varphi_1,\dots,\varphi_{N'^{tr}})$ can obtained as solution to quadratic programming problem by maximization of the dual Lagrangian function. Once $\varphi$ are determined the parameters $w$ and $w_0$ can be determined using $\varphi_m,\,\,\, 1\le m\le N'^{tr}$, respectively, as
\[w=\sum_{m=1}^{N'^{tr}}I(\varphi_m>0)\,\varphi_m\,y^{tr}_m\,X^{tr}_m\]
and
\[w_0=y^{tr}_m-w^T\,X^{tr}_m\,\,\,\mbox{for}\,\,\,0<\varphi_m<\beta,\,\,\,1\le m\le N'^{tr}.\]
Then the output of this two class classifier for the feature vector $X^{ts}_{m_1}$ is
\[y^{p,ts}_{m_1}(X^{ts}_{m_1})=w^T\,X^{ts}_{m_1}+w_0.\]
The decision of this two class classifier on the class label of the feature vector $X^{ts}_{m_1}$ is
\[y^{d,ts}_{m_1}=\begin{cases}1,&\mbox{if}\,\,\,y^{p,ts}_{m_1}(X^{ts}_{m_1})>0\\
2,&\mbox{if}\,\,\,y^{p,ts}_{m_1}(X^{ts}_{m_1})<=0\end{cases}.\]
Since there are $N_{ct}$ classes, one-versus-one approach developed by Platt et al. \cite{daga} is used by considering $\frac{N_{ct}(N_{ct}-1)}{2}$ two class classifiers explained above for classification of character samples. Let $(w^{i,j}_1,\dots,w^{i,j}_{N_{ftr}},w^{i,j}_0),\,\,\,1\le i<j\le N_{ct},$ be the parameters defining $\frac{N_{ct}(N_{ct}-1)}{2}$ two class classifiers trained by considering pair of data sets $(d^{tr}_i,\,d^{tr}_j),\,\,\,1\le i<j\le N_{ct}$. The final decision of this collection of two class classifiers on the class label of the feature vector $X^{ts}_{m_1}$ is done through $N_{ct}-1$ intermediate decisions. Starting from $k=1,\,j=2$, the classifier $(w^{k,j},w^{k,j}_0)$ is chosen to classify the feature vector $X^{ts}_{m_1}$.
\[\mbox{If}\,\,\, ({w^{k,j}}^T\,X^{ts}_{m_1}+w^{k,j}_0)>0,\,\,\,k=k,\,\,\,j=\max(k,j)+1,\,\,\,j\le N_{ct},\]
\[\mbox{if}\,\,\, ({w^{k,j}}^T\,X^{ts}_{m_1}+w^{k,j}_0)\le0,\,\,\,k=j,\,\,\,j=\max(k,j)+1,\,\,\,j\le N_{ct}.\]
At the end of $N_{ct}-1$ such decision steps the value of $k$ is the decision of the collection of two class classifiers $H^{col,SVM}(X^{ts}_{m_1})$ on the feature vector $X^{ts}_{m_1}$. 
Then the decision of SVM classifier on the class label of feature vector $X^{ts}_{m_1}\in\mathcal{R}^{N_{ftr}\times 1},\,\,\,1\le m_1\le N'_{ts}$, is 
\[H^{SVM}_{w,m_1}(k)=\begin{cases}1,&\mbox{if}\,\,\,H^{col,SVM}(X^{ts}_{m_1})=k\\0,&\mbox{otherwise}\end{cases},\,\,\,1\le k\le N_{ct},\,\,\,H^{SVM}_{w,{m_1}}\in\{0,1\}^{N_{ct}}.\]
\[\mbox{The accuracy of classifier is}\,\,
H^{a,SVM}_w=\frac{\sum_{m_1=1}^{N^{ts}}I(H^{SVM}_{w,m_1}=Y^{ts}_{m_1})}{N^{ts}}.\]

\subsection{Sub-unit based (SUB) classifier}
\indent An online handwritten character is generated as a sequence of strokes and can be analysed in terms of geometrically homogeneous structures called sub-units as shown by Sharma \cite{hpod}. A character therefore can be considered as a spatial structure that can be organized in terms of sub-units. Sub-units can be extracted from online handwritten character strokes using sub-unit extraction method developed by Sharma \cite{hpod}. A character can then be represented globally in terms of features extracted from the character as a whole and locally in terms of features extracted from the sub-units.\\
\indent Handwritten character samples have a lot of variations because they are produced  by different individuals writing them at different times. These variations can be captured both at the global character level and local sub-unit level using statistical methods. A character $C_m$ has global representation $X_m^g$ in terms of all the points in the character and has local representation $X_m^l$ in terms of points in the sub-units. The structure of a sub-unit is hidden and is visible through observation of the latent variable $Z_m^l$ corresponding to the structure of the sub-unit. The number of sub-units in a character $C_m$ is $N'_m$ which is the particular value of the random variable $N_m$. A character $C_m$ in a particular character class is represented as \[C_m=(X_m^g,N_m,X_m^l),\,\,\,X_m^l=\{X^l_{m,m_1}\}_{m_1=1}^{N'_m},\,\,\,1\le m\le |d|.\] Here $X_m^g$ has a Gaussian distribution with parameter $(\mu,\Sigma)$. Each sub-unit vector in the set $\{X^l_{m,m_1}\}_{m_1=1}^{N'_m}$ is generated from a mixture of Gaussian distributions with parameters $\{\mu_{m''}^l,\Sigma_{m''}^l\}_{m''=1}^{N_h^{su}},\,\,\, \mbox{where}\,\,\, N_h^{su}$ is the number of possible structures or mixture components that sub-units can have. $N_m$ and $Z_m^l$ have discrete distributions with parameters $\gamma$ and $\eta$, respectively. The likelihood function for the characters of this character class is \[P(X^g,N,X^l/w),\,\,\,w=(\mu,\Sigma,\gamma,\eta,\mu^l,\Sigma^l),\] where $w$ represents the parameter of the model. The complete data likelihood function for these characters with structure of $X^l$ visible is given by
\[P(X^g,N,X^l,Z^l/w)=\prod_{m=1}^{|d|}P(X^g_m,N_m,X^l_m,Z^l_m/w),\] where the characters are assumed to be generated independently.
\[P(X^g,N,X^l,Z^l/w)=\prod_{m=1}^{|d|}P(X^g_m/\mu,\Sigma)P(N_m/\gamma)P(Z^l_m/N_m,\eta)P(X^l_m/Z_m^l,\mu^l,\Sigma^l).\]
Here this particular factorization reveals the assumptions of the model. The global variation does not much affect the local variations that result in changes in the number $N_m$ of sub-units. So $N_m$ is assumed to be independent of $X_m^g$. Once the number of sub-units $N_m$ is known, the latent variable $Z_{m,m_1}^l$ corresponding to the structure of a sub-unit is generated independently given $N_m$ and the sub-unit vector $X_{m,m_1}^l$ is generated independently given $Z_{m,m_1}^l$. This way $N_m$ sub-units are assumed to be generated in the character $C_m$.
\[P(X^g,N,X^l,Z^l/w)=\prod_{m=1}^{|d|}P(X^g_m/\mu,\Sigma)\prod_{m=1}^{|d|}P(N_m/\gamma)\prod_{m=1}^{|d|}P(Z^l_m/N_m,\eta)P(X^l_m/Z_m^l,\mu^l,\Sigma^l).\]
\[\begin{array}{ll}P(X^g,N,X^l,Z^l/w)=&\prod_{m=1}^{|d|}\mathcal{N}(X^g_m/\mu,\Sigma)\prod_{m=1}^{|d|}\prod_{m'=1}^{N^{su}}\left(\gamma_{m'}\right)^{N_{m}^{m'}}\\ \\&\prod_{m=1}^{|d|}\prod_{m'=1}^{N^{su}}\prod_{m_1=1}^{N'_m}\prod_{m''=1}^{N_h^{su}}\left({\eta_{m',m''}}\right)^{(N_m^{m'}\,Z_{m,m_1}^{l,m''})}\\ \\&
\prod_{m=1}^{|d|}\prod_{m_1=1}^{N'_m}\prod_{m''=1}^{N_h^{su}}{\mathcal{N}(X^l_{m,m_1}/\mu_{m''}^l,\Sigma_{m''}^l)}^{Z_{m,m_1}^{l,m''}}.\end{array}\]
\[N_m\,\,\, \mbox{has one-of-}N^{su}\,\,\,\mbox{representation.}\,\,\, N_m=(N_m^1,\dots,N_m^{N^{su}}),\,N_m^{m'}\in\{0,1\},\,\,\sum_{m'=1}^{N^{su}}N_m^{m'}=1,\]\[N'_m=m'\,\,\,\mbox{if}\,\,\, N_m^{m'}=1,\,\,\, 0\le \gamma_{m'}\le 1,\,\,\, \sum_{m'=1}^{N^{su}}\gamma_{m'}=1,\,\,\,\mbox{where}\,\,\, N^{su}\,\,\,\mbox{is the number of sub-units}\]\[\mbox{that characters have.}\,\,\,Z_{m,m_1}^{l}\,\,\,\mbox{has one-of-}N_h^{su}\,\,\, \mbox{representation.}\,\,\, Z_{m,m_1}^{l}=(Z_{m,m_1}^{l,1},\dots,Z_{m,m_1}^{l,N_h^{su}}),\]\[Z_{m,m_1}^{l,m''}\in\{0,1\},\,\,\sum_{m''=1}^{N_h^{su}}Z_{m,m_1}^{l,m''}=1,\,\,\, 0\le \eta_{m',m''}\le 1,\,\,\, \sum_{m''=1}^{N_h^{su}}\eta_{m',m''}=1,\,\,\,1\le m'\le N^{su}.\]
\[\begin{array}{lll}log\,P(X^g,N,X^l,Z^l/w)&=&\sum_{m=1}^{|d|}log\left(\mathcal{N}(X^g_m/\mu,\Sigma)\right)+\sum_{m=1}^{|d|}\sum_{m'=1}^{N^{su}}N_{m}^{m'}log\left(\gamma_{m'}\right)\\ \\&&+\sum_{m=1}^{|d|}\sum_{m'=1}^{N^{su}}\sum_{m_1=1}^{N'_m}\sum_{m''=1}^{N_h^{su}}N_m^{m'} Z_{m,m_1}^{l,m''}log\left(\eta_{m',m''}\right)\\ \\&&+\sum_{m=1}^{|d|}\sum_{m_1=1}^{N'_m}\sum_{m''=1}^{N_h^{su}} Z_{m,m_1}^{l,m''}log\left(\mathcal{N}(X^l_{m,m_1}/\mu_{m''}^l,\Sigma_{m''}^l)\right).\end{array}\]\[\]
\[\begin{array}{ll}log\,P(X^g,N,X^l,Z^l/w)=&\sum_{m=1}^{|d|}log\left(\frac{1}{{2\pi}^{\frac{N_{ftr}}{2}}det(\Sigma)^{\frac{1}{2}}}\exp\left(-\frac{1}{2}(X^g_m-\mu)^T\Sigma^{-1}(X^g_m-\mu)\right)\right)\\ \\ \end{array}\]
\[+\sum_{m=1}^{|d|}\sum_{m'=1}^{N^{su}}N_{m}^{m'}log\left(\gamma_{m'}\right)+\sum_{m=1}^{|d|}\sum_{m'=1}^{N^{su}}\sum_{m_1=1}^{N'_m}\sum_{m''=1}^{N_h^{su}}N_m^{m'} Z_{m,m_1}^{l,m''}log\left(\eta_{m',m''}\right)\]
\[+\sum_{m=1}^{|d|}\sum_{m_1=1}^{N'_m}\sum_{m''=1}^{N_h^{su}} Z_{m,m_1}^{l,m''}log\left(\frac{1}{{2\pi}^{\frac{N^l_{ftr}}{2}}det(\Sigma^l_{m''})^{\frac{1}{2}}}\exp\left(-\frac{1}{2}(X^l_{m,m_1}-\mu_{m''}^l)^T(\Sigma_{m''}^{l})^{-1}(X^l_{m,m_1}-\mu^l_{m''})\right)\right).\]
\[log\,P(X^g,N,X^l,Z^l/w)=\sum_{m=1}^{|d|}\left(-\frac{N_{ftr}}{2}log(2\pi)-\frac{1}{2}log\left(det(\Sigma)\right)-\frac{1}{2}(X^g_m-\mu)^T\Sigma^{-1}(X^g_m-\mu)\right)\]
\[+\sum_{m=1}^{|d|}\sum_{m'=1}^{N^{su}}N_{m}^{m'}log\left(\gamma_{m'}\right)+\sum_{m=1}^{|d|}\sum_{m'=1}^{N^{su}}\sum_{m_1=1}^{N'_m}\sum_{m''=1}^{N_h^{su}}N_m^{m'} Z_{m,m_1}^{l,m''}log\left(\eta_{m',m''}\right)+\sum_{m=1}^{|d|}\sum_{m_1=1}^{N'_m}\sum_{m''=1}^{N_h^{su}}\]
\[Z_{m,m_1}^{l,m''}\bigg(-\frac{N^l_{ftr}}{2}log(2\pi)-\frac{1}{2}log\left(det(\Sigma^l_{m''})\right)-\frac{1}{2}(X^l_{m,m_1}-\mu^l_{m''})^T(\Sigma_{m''}^l)^{-1}(X^l_{m,m_1}-\mu_{m''}^l)\bigg).\]
Maximum likelihood estimation of parameters of the model with latent variables can be done using Expectation Maximization (EM) algorithm. EM algorithm is an iterative process that has two steps per iteration. In the E-step the function 
\[F(w;w^o)=\mathbf{E}_{Z^l/X^g,N,X^l,w^o}\,log\,P(X^g,N,X^l,Z^l/w)\]
is determined. This is the conditional expectation of $log\,P(X^g,N,X^l,Z^l/w)$ with respect to $Z^l$ conditioned on $X^g$, $N$,  $X^l$ and $w^o$ is the estimated parameters in the previous iteration. 
In the M-step $w$ is determined as
\[w=\underset{w'}{arg\,max}\,\, F(w';w^o)\] and this process is repeated till the likelihood function $P(X^g,N,X^l/w)$ converges to a suitable value.
\[F(w;w^o)=\sum_{m=1}^{|d|}\left(-\frac{N_{ftr}}{2}log(2\pi)-\frac{1}{2}log\left(det(\Sigma)\right)-\frac{1}{2}(X^g_m-\mu)^T\Sigma^{-1}(X^g_m-\mu)\right)\]
\[+\sum_{m=1}^{|d|}\sum_{m'=1}^{N^{su}}N_{m}^{m'}log\left(\gamma_{m'}\right)+\sum_{m=1}^{|d|}\sum_{m'=1}^{N^{su}}\sum_{m_1=1}^{N'_m}\sum_{m''=1}^{N_h^{su}}N_m^{m'}\,\,\mathbf{E}_{Z^l/X^g,N,X^l,w^o}\left( Z_{m,m_1}^{l,m''}\right)log\left(\eta_{m',m''}\right)\]\[+\sum_{m=1}^{|d|}\sum_{m_1=1}^{N'_m}\sum_{m''=1}^{N_h^{su}}\mathbf{E}_{Z^l/X^g,N,X^l,w^o}\left(Z_{m,m_1}^{l,m''}\right) \bigg(-\frac{N^l_{ftr}}{2}log(2\pi)-\frac{1}{2}log\left(det(\Sigma^l_{m''})\right)\]\[-\frac{1}{2}(X^l_{m,m_1}-\mu^l_{m''})^T(\Sigma_{m''}^l)^{-1}(X^l_{m,m_1}-\mu_{m''}^l)\bigg).\]
\[\mbox{Let}\,\,\,  \rho(Z_{m,m_1}^{l,m''})=\mathbf{E}_{Z^l/X^g,N,X^l,w^o}\left(Z_{m,m_1}^{l,m''}\right)=Pr(Z_{m,m_1}^{l,m''}=1/X^g,N,X^l,w^o).\]
\[\begin{array}{lll}\rho(Z_{m,m_1}^{l,m''})&=&\frac{P(X^g,N,X^l,Z_{m,m_1}^{l,m''}/w^o)}{\sum_{m'''=1}^{N_h^{su}}P(X^g,N,X^l,Z_{m,m_1}^{l,m'''}/w^o)}\\ \\&=&\frac{P(X^g/w^o)P(N/w^o)P(Z_{m,m_1}^{l,m''}/N,w^o)P(X_{m,m_1}^l/Z_{m,m_1}^{l,m''},w^o)}{\sum_{m'''=1}^{N_h^{su}}P(X^g/w^o)P(N/w^o)P(Z_{m,m_1}^{l,m'''}/N,w^o)P(X_{m,m_1}^l/Z_{m,m_1}^{l,m'''},w^o)}.\end{array}\]\[\rho(Z_{m,m_1}^{l,m''})=\frac{P(Z_{m,m_1}^{l,m''}/N,w^o)P(X_{m,m_1}^l/Z_{m,m_1}^{l,m''},w^o)}{\sum_{m'''=1}^{N_h^{su}}P(Z_{m,m_1}^{l,m'''}/N,w^o)P(X_{m,m_1}^l/Z_{m,m_1}^{l,m'''},w^o)}=\frac{\eta^o_{m',m''}\,\mathcal{N}(X_{m,m_1}^l/\mu_{m''}^{l,o},\Sigma_{m''}^{l,o})}{\sum_{m'''=1}^{N_h^{su}}\eta^o_{m',m'''}\,\mathcal{N}(X_{m,m_1}^{l}/\mu_{m'''}^{l,o},\Sigma_{m'''}^{l,o})}.\]
\[F(w;w^o)=\sum_{m=1}^{|d|}\left(-\frac{N_{ftr}}{2}log(2\pi)-\frac{1}{2}log\left(det(\Sigma)\right)-\frac{1}{2}(X^g_m-\mu)^T\Sigma^{-1}(X^g_m-\mu)\right)+\sum_{m=1}^{|d|}\sum_{m'=1}^{N^{su}}N_{m}^{m'}log\left(\gamma_{m'}\right)\]
\[+\sum_{m=1}^{|d|}\sum_{m'=1}^{N^{su}}\sum_{m_1=1}^{N'_m}\sum_{m''=1}^{N_h^{su}}N_m^{m'}\,\,\rho( Z_{m,m_1}^{l,m''})log\left(\eta_{m',m''}\right)+\sum_{m=1}^{|d|}\sum_{m_1=1}^{N'_m}\sum_{m''=1}^{N_h^{su}} \,\rho( Z_{m,m_1}^{l,m''})\]
\[\bigg(-\frac{N^l_{ftr}}{2}log(2\pi)-\frac{1}{2}log\left(det(\Sigma^l_{m''})\right)-\frac{1}{2}(X^l_{m,m_1}-\mu^l_{m''})^T(\Sigma_{m''}^l)^{-1}(X^l_{m,m_1}-\mu_{m''}^l)\bigg).\]
The values of $w$ for which $F(w;w^o)$ is maximized are obtained by equating the derivatives of $F(w;w^o)$ with respect to the parameters $w=(\mu,\Sigma,\gamma,\eta,\mu^l,\Sigma^l)$ to zero.
\[\frac{\partial F(w;w^o)}{\partial \mu}=0\Longrightarrow\sum_{m=1}^{|d|}-\Sigma^{-1}(X^g_m-\mu)=0\Longrightarrow \mu=\frac{\sum_{m=1}^{|d|}X^g_m}{|d|}.\]
\[\frac{\partial F(w;w^o)}{\partial \Sigma}=0\Longrightarrow -\sum_{m=1}^{|d|}\frac{\partial log\left(det(\Sigma)\right)}{\partial \Sigma}-\sum_{m=1}^{|d|}\frac{\partial \left((X^g_m-\mu)^T\Sigma^{-1}(X^g_m-\mu)\right)}{\partial \Sigma}=0\Longrightarrow \]\[-\sum_{m=1}^{|d|}\frac{det(\Sigma)\,\Sigma^{-1}}{det(\Sigma)}-\sum_{m=1}^{|d|}\left(-\Sigma^{-1}\,(X^g_m-\mu)(X^g_m-\mu)^T\Sigma^{-1}\right)=0\Longrightarrow\Sigma=\frac{\sum_{m=1}^{|d|}(X^g_m-\mu)(X^g_m-\mu)^T}{|d|}.\]
\[\frac{\partial} {\partial \gamma_{m'}}\left(F(w;w^o)+\lambda^{l\gamma}\big(1-\sum_{m'=1}^{N^{su}}\gamma_{m'}\big)\right)=0\Longrightarrow\sum_{m=1}^{|d|}\frac{N_m^{m'}}{\gamma_{m'}}-\lambda^{l\gamma}=0\Longrightarrow \gamma_{m'}= \frac{1}{\lambda^{l\gamma}}\sum_{m=1}^{|d|}{N_m^{m'}}.\]
\[\sum_{m'=1}^{N^{su}}\gamma_{m'}=1,\,\,\,\gamma_{m'}= \frac{1}{\lambda^{l\gamma}}\sum_{m=1}^{|d|}N_m^{m'}\Longrightarrow \gamma_{m'}=\frac{\sum_{m=1}^{|d|}N_m^{m'}}{\sum_{m=1}^{|d|}\sum_{m'=1}^{N^{su}}N_m^{m'}}.\]
$\lambda^{l\gamma}$ is Lagrange multiplier as described in Chong et al. \cite{am} for the constraint $\sum_{m'=1}^{N^{su}}\gamma_{m'}=1$. 
\[\frac{\partial} {\partial \eta_{m',m''}}\left(F(w;w^o)+\lambda^{l\eta}\big(1-\sum_{m''=1}^{N_h^{su}}\eta_{m',m''}\big)\right)=0\]\[\Longrightarrow \sum_{m=1}^{|d|}\sum_{m_1=1}^{N'_m}\frac{N_m^{m'}\rho( Z_{m,m_1}^{l,m''})}{\eta_{m',m''}}-\lambda^{l\eta}=0\Longrightarrow \eta_{m',m''}= \frac{1}{\lambda^{l\eta}}\sum_{m=1}^{|d|}\sum_{m_1=1}^{N'_m}{N_m^{m'}\rho( Z_{m,m_1}^{l,m''})}.\]
\[\sum_{m''=1}^{N_h^{su}}\eta_{m',m''}=1,\,\,\eta_{m',m''}= \frac{1}{\lambda^{l\eta}}\sum_{m=1}^{|d|}\sum_{m_1=1}^{N'_m}N_m^{m'}\rho( Z_{m,m_1}^{l,m''})\Longrightarrow\eta_{m',m''}=\frac{\sum_{m=1}^{|d|}\sum_{m_1=1}^{N'_m}N_m^{m'}\rho( Z_{m,m_1}^{l,m''})}{\sum_{m=1}^{|d|}\sum_{m_1=1}^{N'_m}\sum_{m''=1}^{N_h^{su}}N_m^{m'}\rho( Z_{m,m_1}^{l,m''})}.\]
\[\lambda^{l\eta}\,\,\,\mbox{is Lagrange multiplier for the constraint}\,\,\, \sum_{m''=1}^{N_h^{su}}\eta_{m',m''}=1,\,\,\,1\le m'\le N^{su}.\]
\[\frac{\partial F(w;w^o)}{\partial \mu^l_{m''}}=0\Longrightarrow\sum_{m=1}^{|d|}\sum_{m_1=1}^{N'_m}\left(\rho( Z_{m,m_1}^{l,m''})\,{(\Sigma^l_{m''})}^{-1}(X^l_{m,m_1}-\mu^l_{m''})\right)=0\]
\[\Longrightarrow \mu^l_{m''}=\frac{\sum_{m=1}^{|d|}\sum_{m_1=1}^{N'_m}\rho( Z_{m,m_1}^{l,m''})X^l_{m,m_1}}{\sum_{m=1}^{|d|}\sum_{m_1=1}^{N'_m}\rho( Z_{m,m_1}^{l,m''})},\,\,\,1\le m''\le N_h^{su}.\]
\[\frac{\partial F(w;w^o)}{\partial \Sigma_{m''}^l}=0\Longrightarrow-\sum_{m=1}^{|d|}\sum_{m_1=1}^{N'_m}\rho( Z_{m,m_1}^{l,m''})\frac{\partial}{\partial \Sigma_{m''}^l}\left(log\left(det(\Sigma_{m''}^l)\right)\right)\]
\[-\sum_{m=1}^{|d|}\sum_{m_1=1}^{N'_m}\rho( Z_{m,m_1}^{l,m''})\frac{\partial}{\partial \Sigma^l_{m''}}\left((X^l_{m,m_1}-\mu_{m''}^l)^T(\Sigma_{m''}^l)^{-1}(X^l_{m,m_1}-\mu^l_{m''})\right)=0\]
\[\Longrightarrow -\sum_{m=1}^{|d|}\sum_{m_1=1}^{N'_m}\rho( Z_{m,m_1}^{l,m''})\frac{det(\Sigma_{m''}^l)\,(\Sigma_{m''}^l)^{-1}}{det(\Sigma_{m''}^l)}-\sum_{m=1}^{|d|}\sum_{m_1=1}^{N'_m}\rho( Z_{m,m_1}^{l,m''})\]
\[\left(-(\Sigma_{m''}^l)^{-1}\,(X^l_{m,m_1}-\mu_{m''}^l)(X^l_{m,m_1}-\mu_{m''}^l)^T(\Sigma_{m''}^l)^{-1}\right)=0\]
\[\Longrightarrow \sum_{m=1}^{|d|}\sum_{m_1=1}^{N'_m}\rho( Z_{m,m_1}^{l,m''})(\Sigma_{m''}^l)^{-1}=\sum_{m=1}^{|d|}\sum_{m_1=1}^{N'_m}\rho( Z_{m,m_1}^{l,m''})
\left((\Sigma_{m''}^l)^{-1}\,(X^l_{m,m_1}-\mu_{m''}^l)(X^l_{m,m_1}-\mu_{m''}^l)^T(\Sigma_{m''}^l)^{-1}\right)\]
\[\Longrightarrow\Sigma_{m''}^l=\frac{\sum_{m=1}^{|d|}\sum_{m_1=1}^{N'_m}\rho( Z_{m,m_1}^{l,m''})(X^l_{m,m_1}-\mu_{m''}^l)(X^l_{m,m_1}-\mu_{m''}^l)^T}{\sum_{m=1}^{|d|}\sum_{m_1=1}^{N'_m}\rho( Z_{m,m_1}^{l,m''})},\,\,\,1\le m''\le N_h^{su}.\]\\[15pt]
SUB classifier can be summarized using the algorithm (ASUB) given below.\\[15pt]
\underline{ASUB algorithm}\\
begin Program\\
given character $C_m,\,\,\, 1\le m\le |d|$\\
Compute $X_m^g$, $N'_m$, $X^l_m$, $1\le m\le |d|$\\ 
find $\mu$ and $\Sigma$\\
find $\gamma_{m'},\,1\le m'\le N^{su}$\\
iterate the following steps:\\
(a) find $F(w;w^o)$ for present iteration where $w^o$ has been found in the previous iteration\\
(b) find $\eta_{m',m''},\,1\le m'\le N^{su},\,1\le m''\le N_h^{su}$ and $\mu^l_{m''}$, $\Sigma^l_{m''},\,1\le m''\le N_h^{su}$\\
go to (a)\\
till the likelihood $P(X^g,N,X^l/w)$ has a suitable value.\\
end program\\[15pt]

\indent A Character is represented globally in terms of HPOD features extracted from the whole character region and locally in terms of the HPOD features extracted from the sub-unit regions in the character. Sub-unit extraction method developed by Sharma \cite{hpod} is used to extract sub-units $X^{su,tr}_{k,m,m_1},\,1\le m_1\le N'^{tr}_m,$ from each character $C^{tr}_{k,m}$ of the training set $d_k^{tr}$ of the $k^{th}$ character class. HPOD features $X^{HPOD,tr}_{k,m}$ developed by Sharma \cite{hpod} are extracted from each character $C^{tr}_{k,m}$ and global  feature vector is obtained as  $X^{g,tr}_{k,m}=W^T X^{HPOD,tr}_{k,m}$,  where $W$ is transformation matrix determined using Fisher discriminant method. The local sub-unit feature vector is obtained as  $X^{l,tr}_{k,m,m_1}=[{X^{HPOD,tr}_{k,m,m_1}}^T {X_{k,m,m_1}^{bx,tr}}^T]^T,\,\,1\le m_1\le N'^{tr}_m,$, where $X^{HPOD,tr}_{k,m,m_1}$ is the HPOD feature vector corresponding to $X^{su,tr}_{k,m,m_1}$.
\[X^{bx,tr}_{k,m,m_1}=[min(X^{su,tr}_{k,m,m_1}(*,1))\,\,\,max(X^{su,tr}_{k,m,m_1}(*,1))\,\,\,min(X^{su,tr}_{k,m,m_1}(*,2))\,\,\,max(X^{su,tr}_{k,m,m_1}(*,2))]^T,\]     
$X^{bx,tr}_{k,m,m_1}$ is a vector of minimum and maximum values of x-co-ordinate and y-co-ordinate of points in sub-unit. So  the training set characters $C^{tr}_{k,m}$ are represented as
\[C^{tr}_{k,m}=(X^{g,tr}_{k,m},N'^{tr}_m,X^{l,tr}_{k,m,1},\dots,X^{l,tr}_{k,m,N'^{tr}_m}),\quad 1\le m\le |d_k^{tr}|,\quad 1\le k\le N_{ct}.\] 
SUB method described above is used to estimate the parameters $w_k=(\mu_k,\Sigma_k,\gamma_k,\eta_k,\mu^l_k,\Sigma^l_k),$ from samples  $C^{tr}_{k,m},\,\,1\le m\le |d_k^{tr}|$ of the character class, k, $1\le k\le N_{ct}$. Once the parameters $w=\{w_k\}_{k=1}^{N_{ct}}$ are estimated SUB classifier is used to classify characters from different character classes.\\
The testing set characters $C^{ts}_{k,m}$ are represented similarly as \[C^{ts}_{k,m}=(X^{g,ts}_{k,m},N'^{ts}_m,X^{l,ts}_{k,m,m_1}\dots,X^{l,ts}_{k,m,N'^{ts}_m}),\quad 1\le m\le |d_k^{ts}|,\quad 1\le k\le N_{ct}.\]
If all the testing set character feature samples $C^{ts}_{k,m}$ of all the classes are collected along with their class labels then the testing set can be represented as
\[\{C^{ts}_{m},Y_m^{ts}\}_{m=1}^{N^{ts}},\quad Y^{ts}_m\in\{0,1\}^{N_{ct}},\quad Y^{ts}_m(k)\in\{0,1\},\quad\sum_{k=1}^{N_{ct}}Y^{ts}_m(k)=1,\,\,\,N^{ts}=\sum_{k=1}^{N_{ct}}|d^{ts}_k|.\]
Likelihood of a character $C^{ts}_{m}$ being generated from the $k^{th}$ class is
\[H_{w_k}^{SUB}(C^{ts}_{m})=P(X^{g,ts}_{m},N'^{ts}_m,X^{l,ts}_{m,m_1}\dots,X^{l,ts}_{m,N'^{ts}_m}/w_k),\quad 1\le k\le N_{ct}.\]
The decision of SUB classifier on the class label of character feature sample $C^{ts}_{m},\,\,1\le m\le N^{ts}$, is 
\[H^{SUB}_{w,m}(k)=\begin{cases}1,&\mbox{if}\,\,\,k=\underset{k'}{arg\,max}\,\, H^{SUB}_{w_{k'}}(C^{ts}_m)\\0&\mbox{otherwise}\end{cases},\,\,\,\mbox{for}\,\,\,1\le k,k'\le N_{ct},\,\,\,H^{SUB}_{w,m}\in\{0,1\}^{N_{ct}}.\]
\[\mbox{The accuracy of classifier is}\,\,\,
H^{a,SUB}_w=\frac{\sum_{m=1}^{N^{ts}}I(H^{SUB}_{w,m}=Y^{ts}_m)}{N^{ts}}.\]

\section{Experiments and results}
\indent The dataset consists of samples of 96 different character classes with an average of 133 and 29 samples per character class in the training and testing sets, respectively. Different features used by Sharma \cite{hpod} are extracted from the samples in this dataset. Performances of the classifiers with the feature parameters and the classifier parameters producing the best recognition accuracies on the training and testing sets are presented. Classification performances of the classifiers SOS, SS, FD, FNN and SVM trained with ST, DFT, DCT, DWT, SP, and HOG features with the specific parameter values are given below.
\subsection{Feature parameters}
The size of the feature vectors for different features used by Sharma \cite{hpod} are given below.\\ 
ST features: $N_{ftr}=258$.\\
DFT features: $N_{ftr}=258$.\\
DCT features: $N_{ftr}=258$.\\
DWT features: $N_{ftr}=258$.\\
SP features: $N_{ftr}=786$.\\
HOG features: $N_{ftr}=326$.\\

\subsection{SOS classifier}
Parameters of the classifier are:
$w=(\mu_k,\,\,\,\Sigma_k,\,\,\,1\le k\le 96)$ for each feature type.\\
\begin{table}[ht!]
\caption{Classification accuracies(\%) of SOS classifier on the training set of different feature vectors.}
\begin{center}\begin{tabular}[ht]{|c|c|c|c|c|c|c|}
\hline
&ST&DFT&DCT&DWT&SP&HOG\\
\hline
SOS&88.12&96.67&88.12&88.12&99.62&72.64\\
\hline
\end{tabular}\end{center}\end{table}
\begin{table}[ht!]
\caption{Classification accuracies(\%) of SOS classifier on the testing set of different feature vectors.}
\begin{center}\begin{tabular}[ht]{|c|c|c|c|c|c|c|}
\hline
&ST&DFT&DCT&DWT&SP&HOG\\
\hline
SOS&82.56&82.17&82.56&82.56&64.16&65.72\\
\hline
\end{tabular}\end{center}\end{table}

\subsection{SS classifier}
Parameters of the classifier are:\\
$w=(\{\phi^k_1,\dots,\phi^k_{20}\}_{k=1}^{96},20)$ for ST features\\
$w=(\{\phi^k_1,\dots,\phi^k_{20}\}_{k=1}^{96},20)$ for DFT features\\
$w=(\{\phi^k_1,\dots,\phi^k_{30}\}_{k=1}^{96},30)$ for DCT features\\
$w=(\{\phi^k_1,\dots,\phi^k_{30}\}_{k=1}^{96},30)$ for DWT features\\
$w=(\{\phi^k_1,\dots,\phi^k_{70}\}_{k=1}^{96},70)$ for SP features\\
$w=(\{\phi^k_1,\dots,\phi^k_{70}\}_{k=1}^{96},70)$ for HOG features\\
\begin{table}[ht!]
\caption{Classification accuracies(\%) of SS classifier on the training set of different feature vectors.}
\begin{center}\begin{tabular}[ht]{|c|c|c|c|c|c|c|}
\hline
 &ST&DFT&DCT&DWT&SP&HOG\\
\hline
SS&58.38&58.07&58.52&58.43& 65.20&59.13\\
\hline
\end{tabular}\end{center}\end{table}
\begin{table}[ht!]
\caption{Classification accuracies(\%) of SS classifier on the testing set of different feature vectors.}
\begin{center}\begin{tabular}[ht]{|c|c|c|c|c|c|c|}
\hline
 &ST&DFT&DCT&DWT&SP&HOG\\
\hline
SS&56.58&56.08&56.65&56.58&56.93&56.29\\
\hline
\end{tabular}\end{center}\end{table}

\subsection{FD classifier}
Parameters of the classifier:\\
$w=(\mbox{Fisher transformation matrix}\,\,\,W,\,\,\,\mu^F_k,\,\,\Sigma^F_K,\,\,1\le k\le 96)$.\\
\begin{table}[ht!]
\caption{Classification accuracies(\%) of FD classifier on  the training set of different feature vectors.}
\begin{center}\begin{tabular}[ht]{|c|c|c|c|c|c|c|}
\hline
 &ST&DFT&DCT&DWT&SP&HOG\\
\hline
FD&95.08&97.96&95.08&94.22&98.71&70.77\\
\hline
\end{tabular}\end{center}\end{table}
\begin{table}[ht!]
\caption{Classification accuracies(\%) of FD classifier on the testing set of different feature vectors.}
\begin{center}\begin{tabular}[ht]{|c|c|c|c|c|c|c|}
\hline
 &ST&DFT&DCT&DWT&SP&HOG\\
\hline
FD&86.21&80.86&86.21&84.83&75.36&64.98\\
\hline
\end{tabular}\end{center}\end{table}

\subsection{FNN classifier}
Parameters of the classifier are:\\
$w=(N_{ftr}=258,\,\,N_{hid}=258,\,\,N_{ct}=96)$ for ST features\\
$w=(N_{ftr}=258,\,\,N_{hid}=290,\,\,N_{ct}=96)$ for DFT features\\
$w=(N_{ftr}=258,\,\,N_{hid}=270,\,\,N_{ct}=96)$ for DCT features\\
$w=(N_{ftr}=258,\,\,N_{hid}=270,\,\,N_{ct}=96)$ for DWT features\\
$w=(N_{ftr}=786,\,\,N_{hid}=500,\,\,N_{ct}=96)$ for SP features\\
$w=(N_{ftr}=326,\,\,N_{hid}=524,\,\,N_{ct}=96)$ for HOG features\\
\begin{table}[ht!]
\caption{Classification accuracies(\%) of FNN classifier on the training set of different feature vectors.}
\begin{center}\begin{tabular}[ht]{|c|c|c|c|c|c|c|}
\hline
 &ST&DFT&DCT&DWT&SP&HOG\\
\hline
FNN&96.25&87.32&87.37&94.68&88.58&89.28\\
\hline
\end{tabular}\end{center}\end{table}
\begin{table}[ht!]
\caption{Classification accuracies(\%) of FNN classifier on the testing set of different feature vectors.}
\begin{center}\begin{tabular}[ht]{|c|c|c|c|c|c|c|}
\hline
 &ST&DFT&DCT&DWT&SP&HOG\\
\hline
FNN&86.28&74.41&75.15&85.04&64.74&72.28\\
\hline
\end{tabular}\end{center}\end{table}\newpage

\subsection{SVM classifier}
Parameters of the classifier are:\\
$k^r(X,Y)=\exp(-\frac{1}{\Upsilon^2}||X-Y||^2)$.\\ 
$\beta=1024$ and $\Upsilon=10$ for ST features\\
$\beta=1024$ and $\Upsilon=28$ for DFT features\\
$\beta=1024$ and $\Upsilon=28$ for DCT features\\
$\beta=1024$ and $\Upsilon=20$ for DWT features\\
$\beta=1024$ and $\Upsilon=10$ for SP features\\
$\beta=1024$ and $\Upsilon=10$ for HOG features\\
\begin{table}[ht!]
\caption{Classification accuracies(\%) of SVM classifier on the training set of different feature vectors.}
\begin{center}\begin{tabular}[ht]{|c|c|c|c|c|c|c|}
\hline
 &ST&DFT&DCT&DWT&SP&HOG\\
\hline
SVM&99.9&100&98.27&99.66&100&99.82\\
\hline
\end{tabular}\end{center}\end{table}
\begin{table}[ht!]
\caption{Classification accuracies(\%) of SVM classifier on the testing set of different feature vectors.}
\begin{center}\begin{tabular}[ht!]{|c|c|c|c|c|c|c|}
\hline
 &ST&DFT&DCT&DWT&SP&HOG\\
\hline
SVM&89.22&90.29&86.71&88.34&76.75&77.63\\
\hline
\end{tabular}\end{center}\end{table}

\subsection{Best classification performances of other classifiers on the testing dataset}
\indent SOS classifier has the highest classification accuracy of 82.56\% when trained with ST, DCT and DWT features as given in Table 2. SS classifier has the highest classification accuracy of 56.93\% when trained with SP features as given in Table 4. FD classifier has the highest accuracy of 86.21\% when trained with ST and DCT features as given in Table 6. FNN classifier has the highest accuracy of 86.28\% when trained with ST features as given in Table 8. SVM classifer has the highest accuracy of 90.29\% when trained with DFT features as given in Table 10. From among the other classifiers considered in this study, SVM has the highest classification accuracy of 90.29\% on the testing dataset.   
\subsection{Classification performance of SUB classifier}
\indent SUB classifier derives its stroke direction and stroke order variation independent properties from its use of HPOD features for local and global representations of characters. Classification performance of SUB classifier is compared to that of the other classifiers trained with different features extracted from the same training set and tested on the same testing set.   
\subsubsection{\bfseries{Feature parameters}}
\indent The feature parameter values considered in this experiment have been obtained from Sharma \cite{hpod}. The global HPOD vectors and the local HPOD vectors have sizes of $N_{ftr}=N_{HPOD}=722$ and $N^l_{ftr}=N^l_{HPOD}=134$, respectively.

\subsubsection{\bfseries{Classifier parameters}}
$w=(\mu_k,\Sigma_k,\gamma_k,\eta_k,\mu_k^l,\Sigma_k^l,1\le k\le 96)$. These parameters are estimated using ASUB algorithm.\\
\subsubsection{\bfseries{Classifier accuracy}}
The accuracy of SUB classifier on the training and testing sets are 96.7\% and 93.5\%, respectively.
\subsection{Performance comparison of SUB classifier with that of the other classifiers} 
\indent The classification performance of a classifier is based on its classification accuracy on samples from testing dataset which the classifier has not been trained on. SUB classifier trained with HPOD features has the highest classification accuracy of 93.5\% among all the classifiers considered in this study and evaluated on the same testing dataset. The comparison of classification accuracies of the developed classifier and the other classifiers are shown in the table 11.
\begin{table}[ht!]
\caption{Classification accuracies(\%) of SOS, SS, FD, FNN, SVM classifiers with features giving maximum classification performance and SUB classifier with HPOD features.}
\begin{center}\begin{tabular}[ht!]{|c|c|c|c|c|c|c|}
\hline
 Classifiers&SOS&SS&FD&FNN&SVM&SUB\\
\hline
Features&ST, DCT and DWT&SP&ST and DCT&ST&DFT&HPOD\\
\hline
Classification accuracy&82.56&56.93&86.21&86.28&90.29&93.5\\
\hline
\end{tabular}\end{center}\end{table}

\section{Conclusion}

Different classifiers used for character recognition in other studies are considered in this study for performance comparison with the developed classifier. These classifiers are SOS, SS, FD, FNN and SVM classifiers. These classifiers are trained with ST, DFT, DCT, DWT, SP and HOG features. These features do not explicitly capture local variations. Classifiers trained with ST, DFT, DCT, and DWT features are dependent on stroke direction and order variations. Classifiers trained with SP and HOG features do not have such dependence but still are unable to capture local variations in a handwritten character structure. The sub-unit based (SUB) classifier developed in this work uses HPOD features to represent characters at global character level and local sub-unit level and is robust to local and global variations in the handwritten character structure. This classifier is indepenent of the stroke direction and order variations. SUB classifier has the highest classification accuracy among all the classifiers considered in this study.



\begin{thebibliography}{20}


\ifx \showCODEN    \undefined \def \showCODEN     #1{\unskip}     \fi
\ifx \showDOI      \undefined \def \showDOI       #1{#1}\fi
\ifx \showISBNx    \undefined \def \showISBNx     #1{\unskip}     \fi
\ifx \showISBNxiii \undefined \def \showISBNxiii  #1{\unskip}     \fi
\ifx \showISSN     \undefined \def \showISSN      #1{\unskip}     \fi
\ifx \showLCCN     \undefined \def \showLCCN      #1{\unskip}     \fi
\ifx \shownote     \undefined \def \shownote      #1{#1}          \fi
\ifx \showarticletitle \undefined \def \showarticletitle #1{#1}   \fi
\ifx \showURL      \undefined \def \showURL       {\relax}        \fi
\providecommand\bibfield[2]{#2}
\providecommand\bibinfo[2]{#2}
\providecommand\natexlab[1]{#1}
\providecommand\showeprint[2][]{arXiv:#2}

\bibitem[Aparna et~al\mbox{.}(2004)]%
        {clse}
\bibfield{author}{\bibinfo{person}{K.~H. Aparna}, \bibinfo{person}{Subramanian
  V}, \bibinfo{person}{M. Kasirajan}, \bibinfo{person}{G.~V. Prakash},
  \bibinfo{person}{V.~S. Chakravarthy}, {and} \bibinfo{person}{S. Madhvanath}.}
  \bibinfo{year}{2004}\natexlab{}.
\newblock \showarticletitle{Online handwriting recognition for Tamil}. In
  \bibinfo{booktitle}{\emph{IWFHR}}.
\newblock


\bibitem[Belhe et~al\mbox{.}(2012)]%
        {troaw}
\bibfield{author}{\bibinfo{person}{S. Belhe}, \bibinfo{person}{C. Paulzagade},
  \bibinfo{person}{A. Deshukh}, \bibinfo{person}{S. Jetley}, {and}
  \bibinfo{person}{K. Mehrotra}.} \bibinfo{year}{2012}\natexlab{}.
\newblock \showarticletitle{Hindi handwritten word recognition using HMM and
  symbol tree}. In \bibinfo{booktitle}{\emph{DAR}}.
\newblock


\bibitem[Chong and Zak(2001)]%
        {am}
\bibfield{author}{\bibinfo{person}{E.~K.~P. Chong} {and} \bibinfo{person}{S.~H.
  Zak}.} \bibinfo{year}{2001}\natexlab{}.
\newblock \bibinfo{booktitle}{\emph{An introduction to optimization}}.
\newblock \bibinfo{publisher}{Wiley}.
\newblock


\bibitem[Connell et~al\mbox{.}(2000)]%
        {troan}
\bibfield{author}{\bibinfo{person}{S.~D. Connell}, \bibinfo{person}{R.~M.~K.
  Sinha}, {and} \bibinfo{person}{A.~K. Jain}.} \bibinfo{year}{2000}\natexlab{}.
\newblock \showarticletitle{Recognition of unconstrained on-line devanagari
  characters}. In \bibinfo{booktitle}{\emph{ICPR}}.
\newblock


\bibitem[Cortes and Vapnik(1995)]%
        {clsfff}
\bibfield{author}{\bibinfo{person}{C. Cortes} {and} \bibinfo{person}{V.
  Vapnik}.} \bibinfo{year}{1995}\natexlab{}.
\newblock \showarticletitle{Support-vector networks}. In
  \bibinfo{booktitle}{\emph{ML}}.
\newblock


\bibitem[Duda et~al\mbox{.}(2006)]%
        {troam}
\bibfield{author}{\bibinfo{person}{R.~O. Duda}, \bibinfo{person}{P.~E. Hart},
  {and} \bibinfo{person}{D.~G. Stork}.} \bibinfo{year}{2006}\natexlab{}.
\newblock \bibinfo{booktitle}{\emph{Pattern classification}}.
\newblock \bibinfo{publisher}{Wiley}.
\newblock


\bibitem[Haykin(1999)]%
        {troak}
\bibfield{author}{\bibinfo{person}{S. Haykin}.}
  \bibinfo{year}{1999}\natexlab{}.
\newblock \bibinfo{booktitle}{\emph{Neural network: a comprehensive
  foundation}}.
\newblock \bibinfo{publisher}{Pearson Education}.
\newblock


\bibitem[Joshi et~al\mbox{.}(2005)]%
        {troao}
\bibfield{author}{\bibinfo{person}{N. Joshi}, \bibinfo{person}{G. Sita},
  \bibinfo{person}{A.~G. Ramakrishnan}, \bibinfo{person}{V. Deepu}, {and}
  \bibinfo{person}{S. Madhvanath}.} \bibinfo{year}{2005}\natexlab{}.
\newblock \showarticletitle{Machine recognition of online handwritten
  Devanagari characters}. In \bibinfo{booktitle}{\emph{ICDAR}}.
\newblock


\bibitem[Joshi et~al\mbox{.}(2004)]%
        {clsa}
\bibfield{author}{\bibinfo{person}{N. Joshi}, \bibinfo{person}{G. Sita},
  \bibinfo{person}{A.~G. Ramakrishnan}, {and} \bibinfo{person}{S. Madhvanath}.}
  \bibinfo{year}{2004}\natexlab{}.
\newblock \showarticletitle{Comparison of elastic matching algorithms for
  online Tamil handwritten character recognition}. In
  \bibinfo{booktitle}{\emph{IWFHR}}.
\newblock


\bibitem[Kubatur et~al\mbox{.}(2012)]%
        {troav}
\bibfield{author}{\bibinfo{person}{S. Kubatur}, \bibinfo{person}{M.~S. Ahmed},
  {and} \bibinfo{person}{M. Ahmadi}.} \bibinfo{year}{2012}\natexlab{}.
\newblock \showarticletitle{A neural network approach to online Devanagari
  handwritten character recognition}. In \bibinfo{booktitle}{\emph{HPCS}}.
\newblock


\bibitem[Lajish and Kopparapu(2010)]%
        {troau}
\bibfield{author}{\bibinfo{person}{V.~L. Lajish} {and} \bibinfo{person}{S.~K.
  Kopparapu}.} \bibinfo{year}{2010}\natexlab{}.
\newblock \showarticletitle{Fuzzy directional features for unconstrained
  on-line Devanagari handwriting recognition}. In
  \bibinfo{booktitle}{\emph{NCC}}.
\newblock


\bibitem[Mehrotra et~al\mbox{.}(2013)]%
        {troax}
\bibfield{author}{\bibinfo{person}{K. Mehrotra}, \bibinfo{person}{S. Jetley},
  \bibinfo{person}{A. Deshmukh}, {and} \bibinfo{person}{S. Belhe}.}
  \bibinfo{year}{2013}\natexlab{}.
\newblock \showarticletitle{Unconstrained handwritten Devanagari character
  recognition using convolutional neural networks}. In
  \bibinfo{booktitle}{\emph{MOCR}}.
\newblock


\bibitem[of~Indian Standard~(BIS)(1991)]%
        {troiscii}
\bibfield{author}{\bibinfo{person}{Bureau of Indian Standard~(BIS)}.}
  \bibinfo{year}{1991}\natexlab{}.
\newblock \bibinfo{title}{Indian script code for information interhange
  (ISCII)}.
\newblock
\newblock


\bibitem[Parui et~al\mbox{.}(2008)]%
        {clsc}
\bibfield{author}{\bibinfo{person}{S.~K. Parui}, \bibinfo{person}{K. Guin},
  \bibinfo{person}{U. Bhattacharya}, {and} \bibinfo{person}{B.~B. Chaudhuri}.}
  \bibinfo{year}{2008}\natexlab{}.
\newblock \showarticletitle{Online handwritten Bangla character recognition
  using HMM}. In \bibinfo{booktitle}{\emph{ICPR}}.
\newblock


\bibitem[Platt et~al\mbox{.}(2000)]%
        {daga}
\bibfield{author}{\bibinfo{person}{J. Platt}, \bibinfo{person}{N. Cristianini},
  {and} \bibinfo{person}{J. Shawe-Taylor}.} \bibinfo{year}{2000}\natexlab{}.
\newblock \showarticletitle{Large margin DAGs for multiclass classification}.
  In \bibinfo{booktitle}{\emph{Advances in Neural Information Processing
  Systems}}.
\newblock


\bibitem[Prasad et~al\mbox{.}(2010)]%
        {clsb}
\bibfield{author}{\bibinfo{person}{M.M. Prasad}, \bibinfo{person}{M. Sukumar},
  {and} \bibinfo{person}{A.~G. Ramakrishnan}.} \bibinfo{year}{2010}\natexlab{}.
\newblock \showarticletitle{Orthogonal LDA in PCA transformed subspace}. In
  \bibinfo{booktitle}{\emph{ICFHR}}.
\newblock


\bibitem[Sharma(2019)]%
        {hpod}
\bibfield{author}{\bibinfo{person}{Anand Sharma}.}
  \bibinfo{year}{2019}\natexlab{}.
\newblock \emph{\bibinfo{title}{Devanagari Online Handwritten Character
  Recognition}}.
\newblock \bibinfo{thesistype}{Ph.\,D. Dissertation}. \bibinfo{school}{Indian
  Institute of Science, Bangaluru, Karnataka, India}.
\newblock


\bibitem[Sharma et~al\mbox{.}(2008)]%
        {clsd}
\bibfield{author}{\bibinfo{person}{A. Sharma}, \bibinfo{person}{R. Kumar},
  {and} \bibinfo{person}{R.~K. Sharma}.} \bibinfo{year}{2008}\natexlab{}.
\newblock \showarticletitle{Online handwritten Gurmukhi character recognition
  using elastic matching}. In \bibinfo{booktitle}{\emph{CISP}}.
\newblock


\bibitem[Swethalakshmi et~al\mbox{.}(2006)]%
        {troaq}
\bibfield{author}{\bibinfo{person}{H. Swethalakshmi}, \bibinfo{person}{A.
  Jayaraman}, \bibinfo{person}{V.~S. Chakravarthy}, {and}
  \bibinfo{person}{C.~C. Sekhar}.} \bibinfo{year}{2006}\natexlab{}.
\newblock \showarticletitle{Online handwritten character recognition of
  Devanagari and Telugu characters using support vector machines}. In
  \bibinfo{booktitle}{\emph{IWFHR}}.
\newblock


\bibitem[toolkit(2014)]%
        {trohp}
\bibfield{author}{\bibinfo{person}{Lipi toolkit}.}
  \bibinfo{year}{2014}\natexlab{}.
\newblock \bibinfo{booktitle}{\emph{HP labs India Indic handwriting datasets}}.
\newblock
\urldef\tempurl%
\url{http://lipitk.sourceforge.net/datasets/ dvngchardata.htm}
\showURL{%
\tempurl}


\end{thebibliography}
\end{document}